%% file: main.tex
\documentclass[10pt,twocolumn,letterpaper]{article}

\usepackage[pagenumbers]{cvpr} 

\usepackage{amsmath}
\usepackage{amssymb}
\usepackage{booktabs}
\usepackage{multirow} 
\usepackage[figuresright]{rotating}
\usepackage{color}
\usepackage{tabularx}
\usepackage[accsupp]{axessibility} 

\input{preamble}
\definecolor{cvprblue}{rgb}{0.21,0.49,0.74}
\usepackage[pagebackref,breaklinks,colorlinks,allcolors=cvprblue]{hyperref}

\def\paperID{5052} 
\def\confName{CVPR}
\def\confYear{2026}

\title{AdaIAT: Adaptively Increasing Attention to Generated Text to Alleviate Hallucinations in LVLM}

\author{
    Li'an Zhong\textsuperscript{1},
    Ziqiang He\textsuperscript{1},
    Jibin Zheng\textsuperscript{2},
    Jin Li\textsuperscript{1},
    Z. Jane Wang\textsuperscript{3},
    Xiangui Kang\textsuperscript{1}\thanks{Corresponding author.} \\
    \textsuperscript{1}Sun Yat-Sen University,
    \textsuperscript{2}Foshan University,
    \textsuperscript{3}University of British Columbia \\
}

\begin{document}
\maketitle
\input{sec/0_abstract}    
\input{sec/1_intro}

\input{sec/2_relatedWork}

\input{sec/3_method}
\input{sec/4_exp}
{
    \small
    \bibliographystyle{ieeenat_fullname}
    \bibliography{main}
}


\end{document}


\maketitle

\begin{table*}[htbp]
  \centering
  \caption{Comparison result of hallucination mitigation and time cost performance of different methods on LLaVA-1.5-7B}
    \begin{tabular}{cccccccccc}
    \toprule
         Method & Greedy & VCD   & AGLA  & OPERA & LURE  & PAI   & HGAI  & IAT   & AdaIAT \\
    \midrule
    $C_S \downarrow$    & 49.0  & 45.0  & 44.2  & 49.2  & 19.8  & 31.8  & 31.4  & 29.8  & 31.4  \\
    $C_I \downarrow$    & 13.3  & 11.8  & 11.6  & 12.6  & 6.5   & 7.8   & 6.9   & 9.0   & 8.3  \\
    F1 $\uparrow$    & 77.9  & 78.3  & 78.6  & 77.8  & 74.1  & 77.7  & 78.3  & 76.8  & 79.4  \\
    $t(ms/token) \downarrow$ & 109.8  & 307.4  & 415.0  & 274.4  & 916.0  & 114.5  & 112.4  & 111.5  & 114.0  \\
    \bottomrule
    \end{tabular}%
  \label{tab:comparison}%
\end{table*}%

\section{Observation and Analysis}

In this section, we conduct additional experiments on LLaVA-1.5-13B~\cite{llava1.5}, Janus-Pro-7B~\cite{chen2025janus}, and Qwen2.5-VL-7B~\cite{bai2025qwen} to verify that the similar characteristics observed for LLaVA‑1.5‑7B also hold for other LVLMs.

\subsection{Attention Pattern of Different LVLMs}
\begin{figure*}[htbp]
\centering
\subfloat[$\bar{\mathbf{A}}_{T_p}^{r}$ and $\bar{\mathbf{A}}_{T_p}^{h}$ (Layers 5-18)]{
    \centering
    \includegraphics[width=0.5\linewidth]{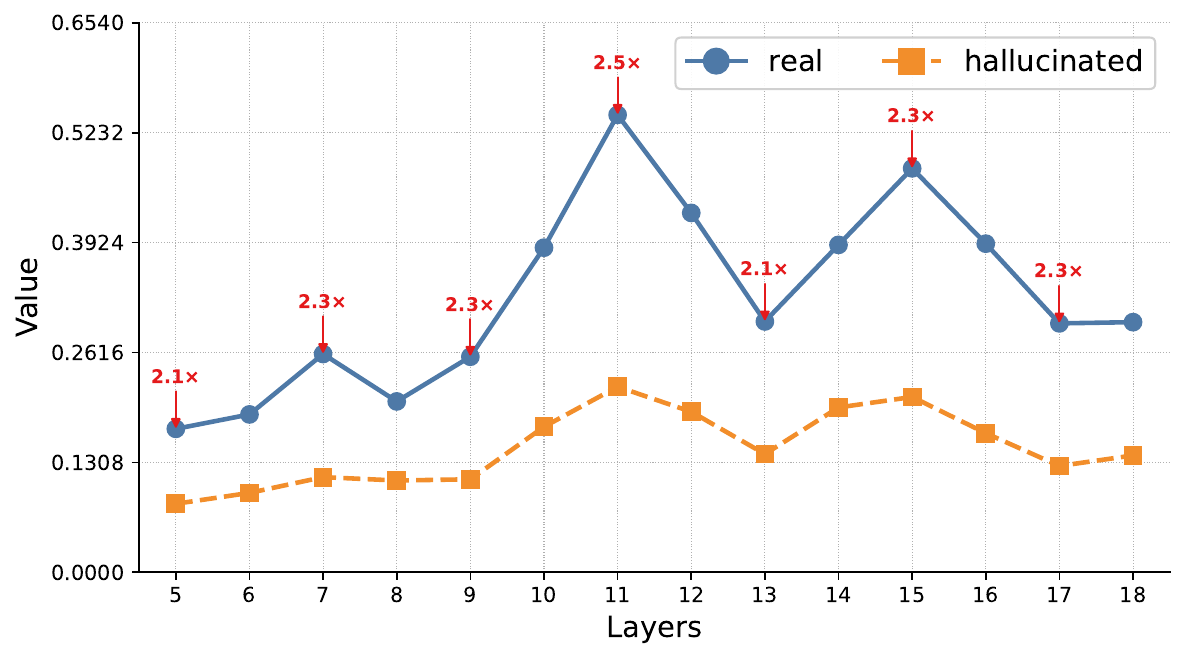}
    \label{fig:contrast_text}
}
\subfloat[$\bar{\mathbf{A}}_{V}^{r}$ and $\bar{\mathbf{A}}_{V}^{h}$ (Layers 5-18)]{
    \centering
    \includegraphics[width=0.5\linewidth]{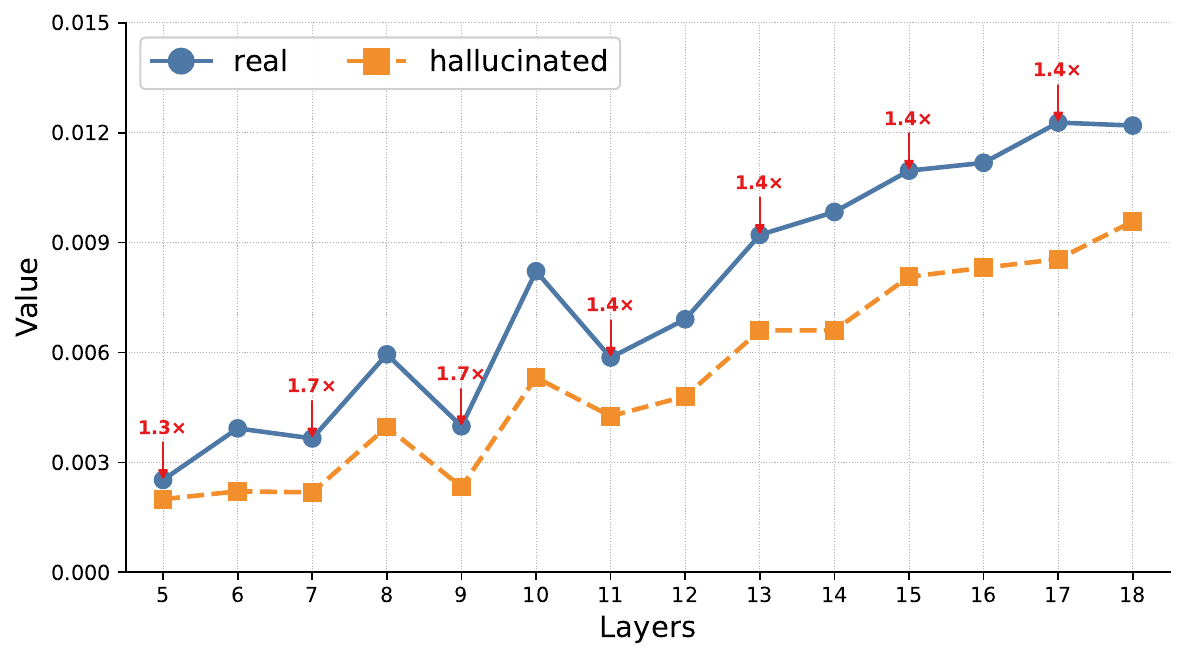}
    \label{fig:contrast_img}
}
\caption{Visualization of the average per-token attention weights from $t_{n+1}$ to $T_{p}$ ($\bar{\mathbf{A}}_{T_p}^{r}$ and $\bar{\mathbf{A}}_{T_p}^{h}$) and to $V$ ($\bar{\mathbf{A}}_{V}^{r}$ and $\bar{\mathbf{A}}_{V}^{h}$) for LLaVA-1.5-13B, showing only layers 5–18 for clearer observation.}
\label{fig:contrast_llava}
\end{figure*}

\begin{figure*}[htbp]
\centering
\subfloat[$\bar{\mathbf{A}}_{T_p}^{r}$ and $\bar{\mathbf{A}}_{T_p}^{h}$ (Layers 5-18)]{
    \centering
    \includegraphics[width=0.5\linewidth]{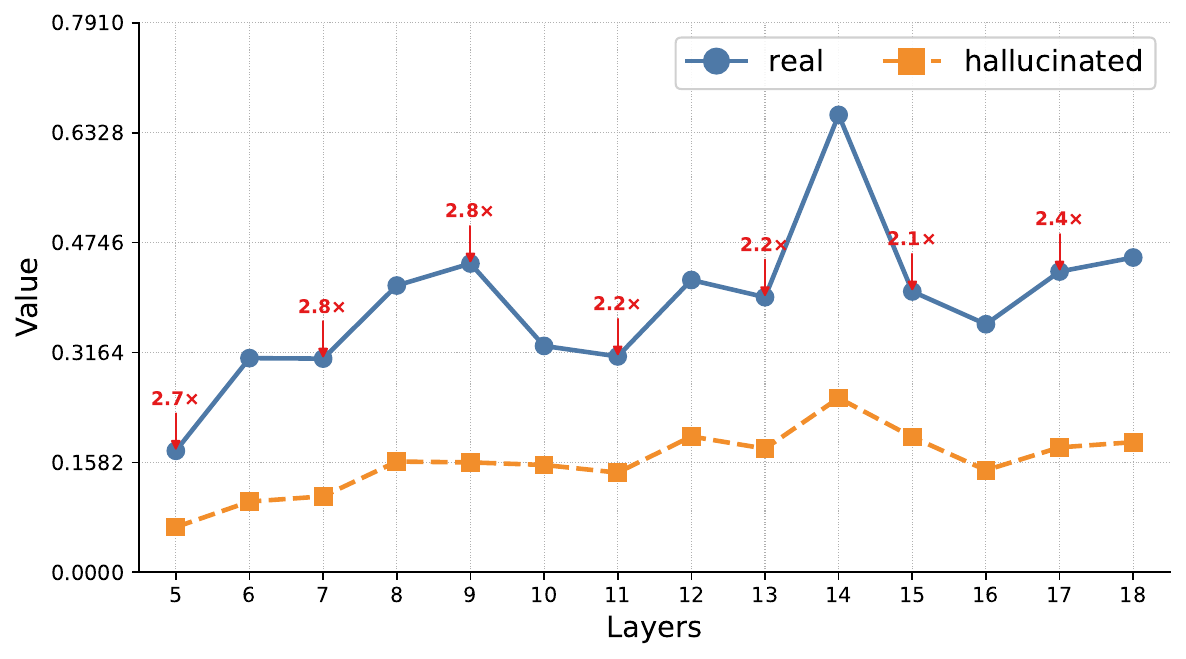}
    \label{fig:contrast_text}
}
\subfloat[$\bar{\mathbf{A}}_{V}^{r}$ and $\bar{\mathbf{A}}_{V}^{h}$ (Layers 5-18)]{
    \centering
    \includegraphics[width=0.5\linewidth]{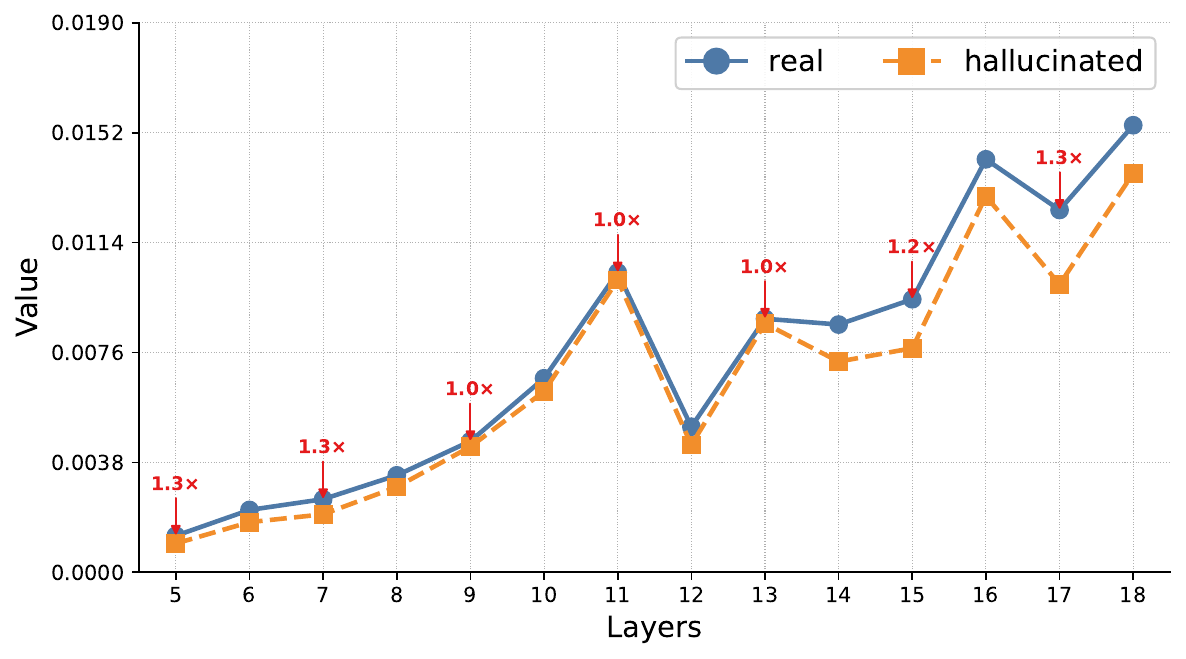}
    \label{fig:contrast_img}
}
\caption{Visualization of the average per-token attention weights from $t_{n+1}$ to $T_{p}$ ($\bar{\mathbf{A}}_{T_p}^{r}$ and $\bar{\mathbf{A}}_{T_p}^{h}$) and to $V$ ($\bar{\mathbf{A}}_{V}^{r}$ and $\bar{\mathbf{A}}_{V}^{h}$) for Janus-Pro-7B, showing only layers 5–18 for clearer observation.}
\label{fig:contrast_janus}
\end{figure*}

\begin{figure*}[htbp]
\centering
\subfloat[$\bar{\mathbf{A}}_{T_p}^{r}$ and $\bar{\mathbf{A}}_{T_p}^{h}$ (Layers 5-18)]{
    \centering
    \includegraphics[width=0.5\linewidth]{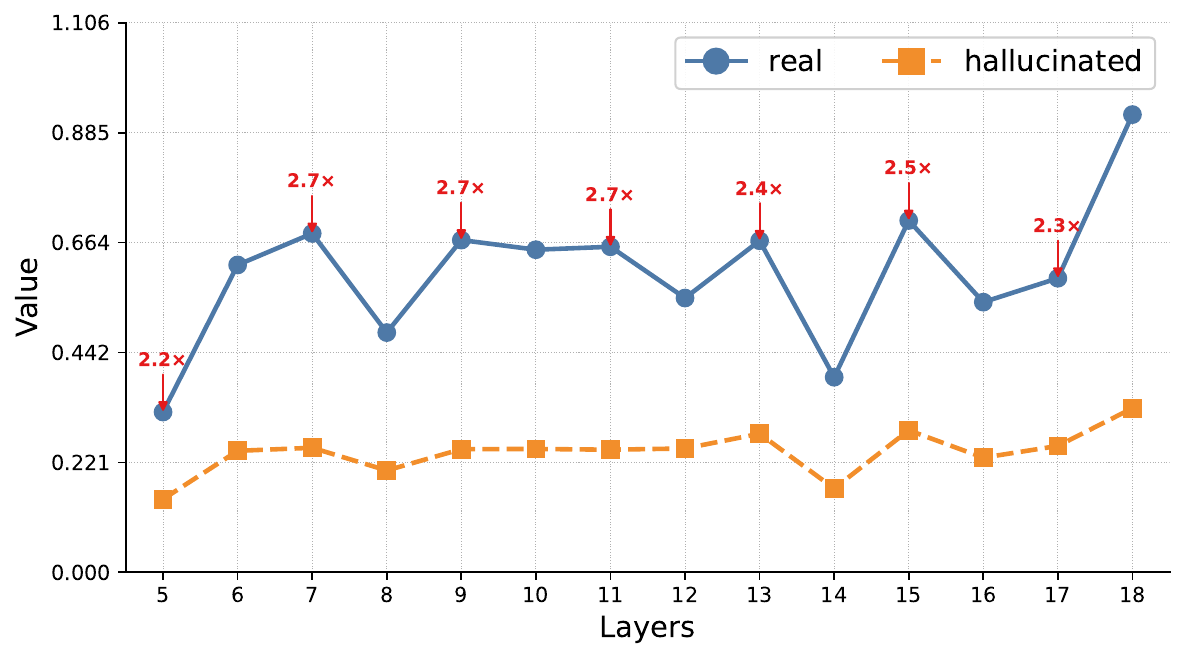}
    \label{fig:contrast_text}
}
\subfloat[$\bar{\mathbf{A}}_{V}^{r}$ and $\bar{\mathbf{A}}_{V}^{h}$ (Layers 5-18)]{
    \centering
    \includegraphics[width=0.5\linewidth]{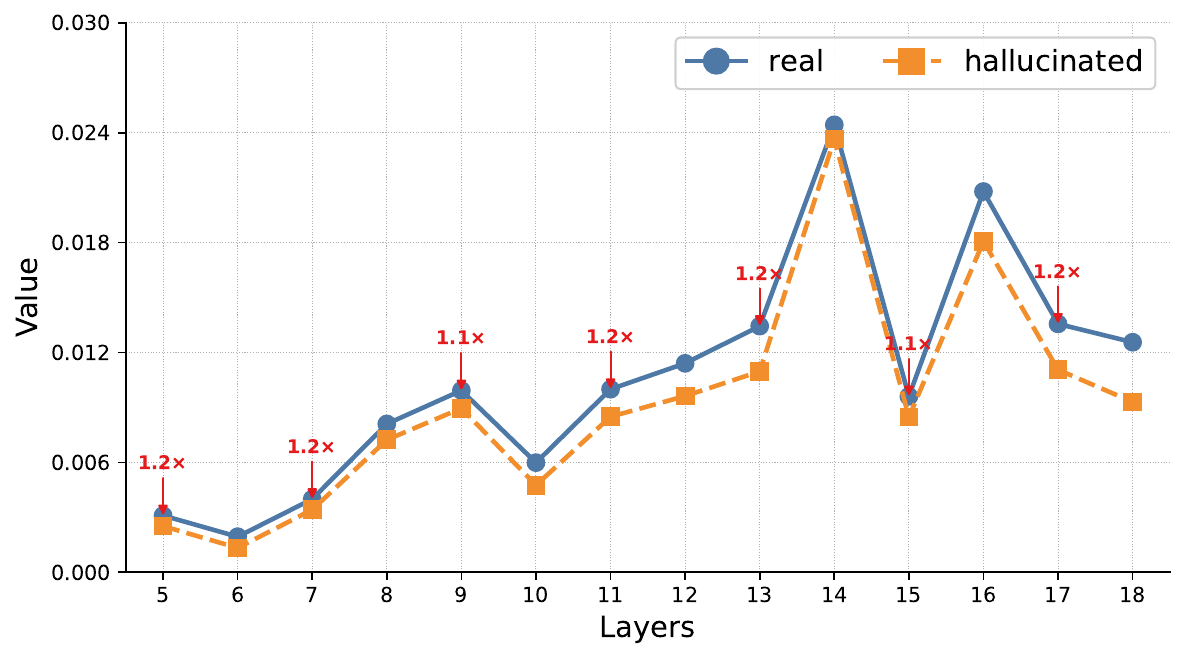}
    \label{fig:contrast_img}
}
\caption{Visualization of the average per-token attention weights from $t_{n+1}$ to $T_{p}$ ($\bar{\mathbf{A}}_{T_p}^{r}$ and $\bar{\mathbf{A}}_{T_p}^{h}$) and to $V$ ($\bar{\mathbf{A}}_{V}^{r}$ and $\bar{\mathbf{A}}_{V}^{h}$) for Qwen2.5-VL-7B, showing only layers 5–18 for clearer observation.}
\label{fig:contrast_qwen}
\end{figure*}

We randomly selected 10,000 images from COCO2014~\cite{coco} and generated captions for these images using each of the three models with the prompt “Please describe the image in detail.” 
Objects in the captions were annotated via ground truth annotations, yielding 22,362 true and 8,978 hallucinated objects for LLaVA-1.5-13B, 15,018 true and 2,924 hallucinated objects for Janus-Pro-7B, and 12,289 true and 2,687 hallucinated objects for Qwen2.5-VL-7B. 
We further computed each model's $\bar{\mathbf{A}}_{T_p}^{r}$, $\bar{\mathbf{A}}_{T_p}^{h}$, $\bar{\mathbf{A}}_{V}^{r}$, and $\bar{\mathbf{A}}_{V}^{h}$, visualized the results in Fig.~\ref{fig:contrast_llava}-\ref{fig:contrast_qwen}.

The experimental results from all three models align with those from LLaVA-1.5-7B: during generation, true objects consistently assign higher average attention to both $V$ and $T_{p}$ than hallucinated objects.
However, the disparity in attention over text tokens (typically on the order of 2x–3×) is substantially greater than that over image tokens (typically on the order of 1x–2×). This observation further supports that the knowledge compressed in $T_{\mathrm{p}}$ may facilitate more accurate predictions.

\subsection{Heads Behavior of Different LVLMs}
\begin{figure}[htbp]
    \centering
    \includegraphics[width=\linewidth]{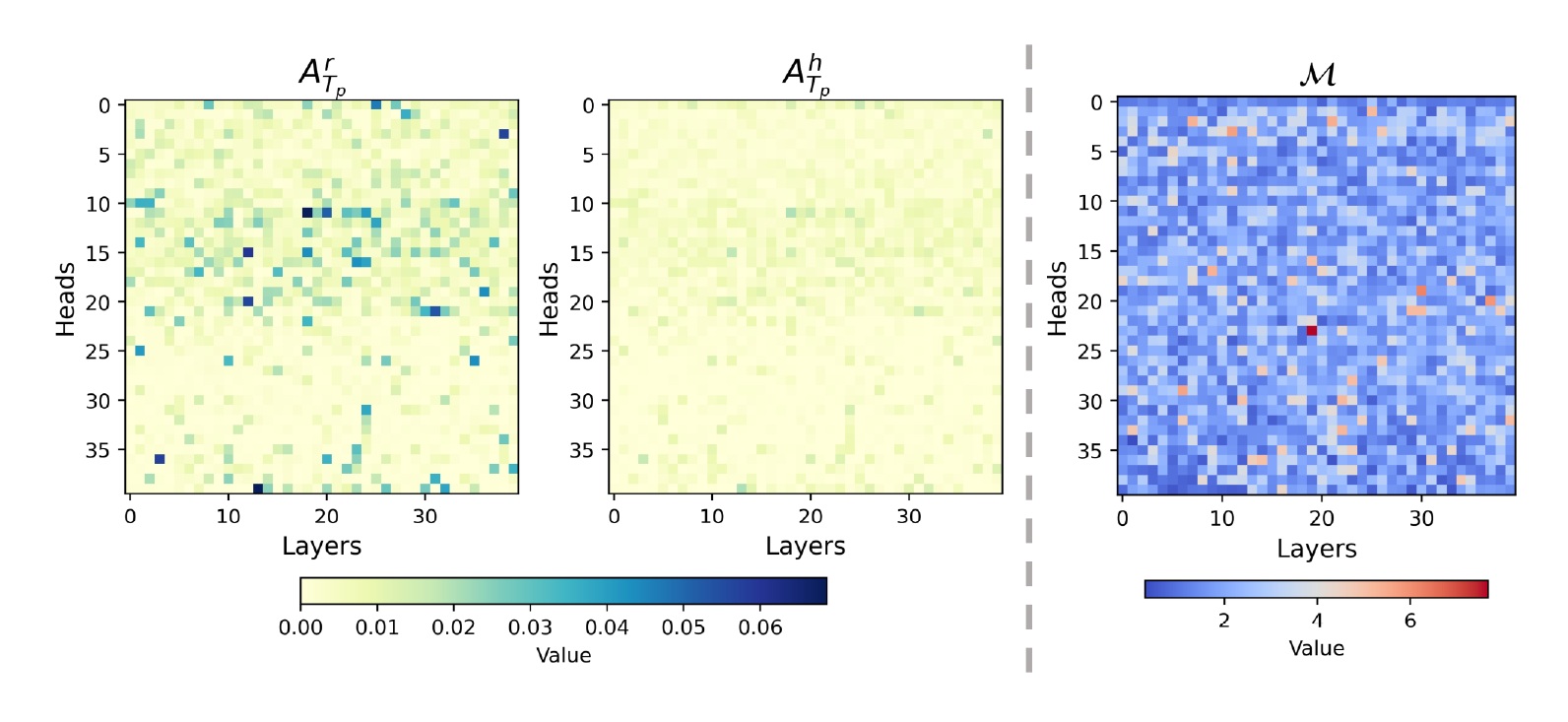}
    \caption{Left: Visualization of the average per‑token attention weight to generated text for each layer and head of the LLaVA-1.5-13B, computed over the sets of real objects ($\bar{\mathbf{A}}_{T_p}^{r}$) and hallucinated objects ($\bar{\mathbf{A}}_{T_p}^{h}$). Right: The visualization of ratio $\bar{\mathbf{A}}_{T_p}^{r} / \bar{\mathbf{A}}_{T_p}^{h}$.}
    \label{fig:heatmap_llava}
\end{figure}

\begin{figure}[htbp]
    \centering
    \includegraphics[width=\linewidth]{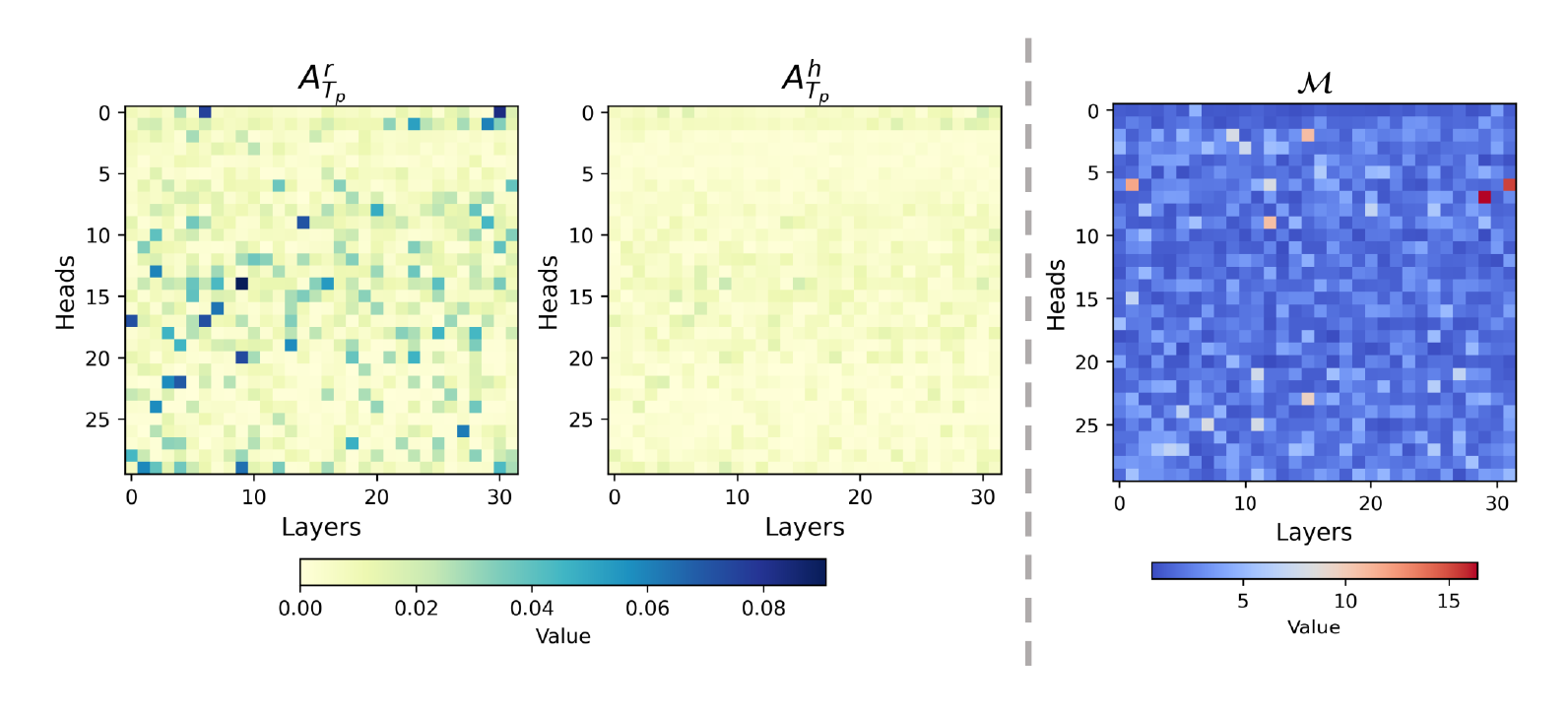}
    \caption{Left: Visualization of the average per‑token attention weight to generated text for each layer and head of the Janus-Pro-7B, computed over the sets of real objects ($\bar{\mathbf{A}}_{T_p}^{r}$) and hallucinated objects ($\bar{\mathbf{A}}_{T_p}^{h}$). Right: The visualization of ratio $\bar{\mathbf{A}}_{T_p}^{r} / \bar{\mathbf{A}}_{T_p}^{h}$.}
    \label{fig:heatmap_janus}
\end{figure}

\begin{figure}[htbp]
    \centering
    \includegraphics[width=\linewidth]{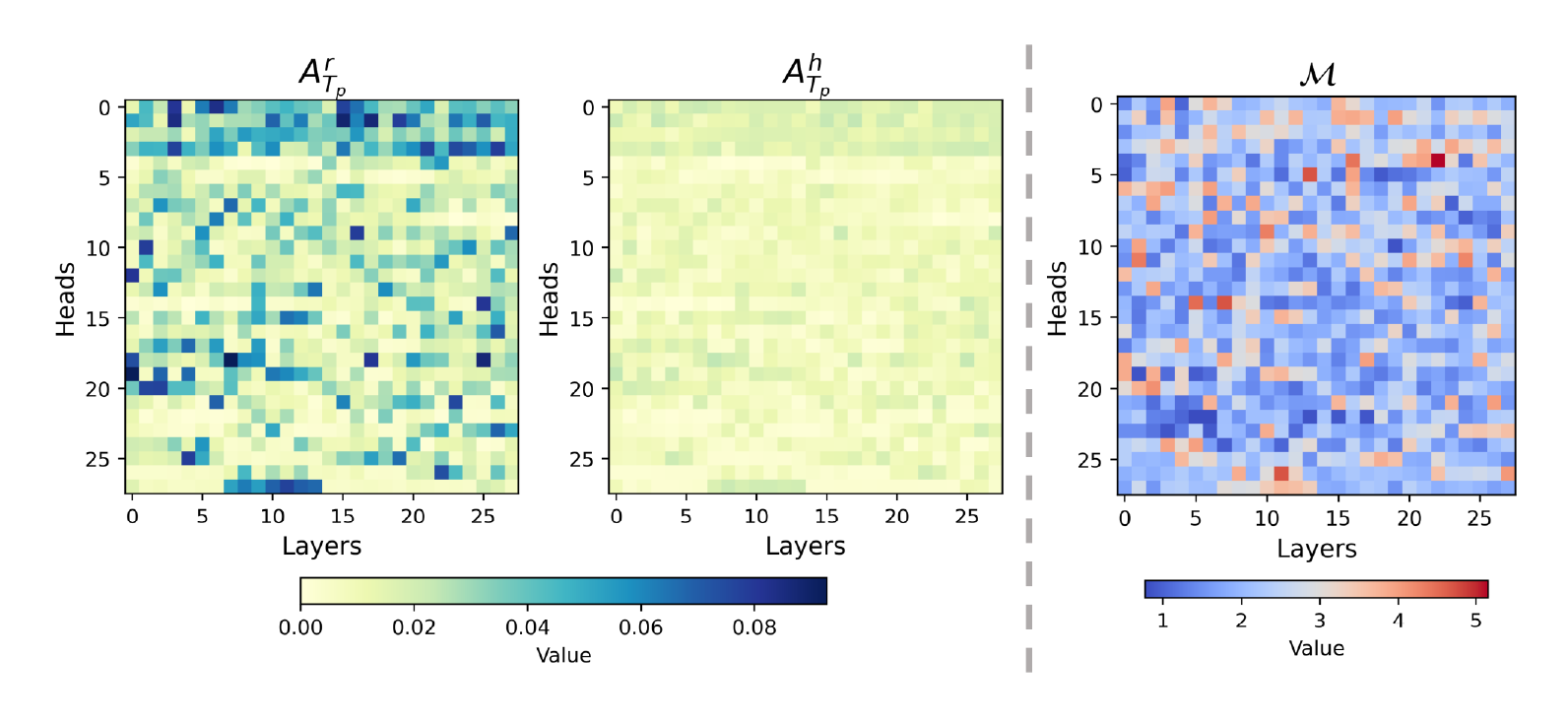}
    \caption{Left: Visualization of the average per‑token attention weight to generated text for each layer and head of the Qwen2.5-VL-7B, computed over the sets of real objects ($\bar{\mathbf{A}}_{T_p}^{r}$) and hallucinated objects ($\bar{\mathbf{A}}_{T_p}^{h}$). Right: The visualization of ratio $\bar{\mathbf{A}}_{T_p}^{r} / \bar{\mathbf{A}}_{T_p}^{h}$.}
    \label{fig:heatmap_qwen}
\end{figure}

For each of the three models (LLaVA-1.5-13B, Janus-Pro-7B, and Qwen2.5-VL-7B), we further computed their $\bar{\mathbf{A}}_{T_p}^{r}$, $\bar{\mathbf{A}}_{T_p}^{h}$, and $\mathcal{M}$, visualizing them in Fig.~\ref{fig:heatmap_llava}-\ref{fig:heatmap_qwen}. Although $\bar{\mathbf{A}}_{T_p}^{r}$ consistently exceeds $\bar{\mathbf{A}}_{T_p}^{h}$, the magnitude of this ratio varies across different attention heads. For LLaVA-1.5-13B and Qwen2.5-VL-7B, some heads exhibit a 2× difference while others reach 5×; For Janus-Pro-7B, certain heads even display a 15× disparity. 
The inherent behavioral differences among attention heads highlight their distinct functional roles. Motivated by this observation, we propose AdaIAT to adaptively increase attention based on each head's specific characteristics.

\section{Extended Experiment results}
\subsection{Comparison with More Methods}

In this section, we compare attention intervention methods (PAI\cite{liu2024pai}, HGAI\cite{devil}, IAT, AdaIAT) with more methods (VCD\cite{leng2024vcd}, AGLA\cite{an2025agla}, LURE\cite{lure}, OPERA\cite{huang2024opera}) on LLaVA-1.5-7B, evaluating their performance in mitigating hallucinations using the CHAIR metric, as well as their corresponding computational costs $t$. The experimental results are shown in the Table~\ref{tab:comparison}. It can be observed that VCD and AGLA, which involve contrastive decoding (each token generation requires two forward passes along with special processing of the input image), incur a time cost 3–4 times that of Greedy decoding. Similarly, OPERA, due to its rollback strategy, results in a time cost more than double that of Greedy decoding. In contrast, attention intervention methods achieve significantly better hallucination reduction while maintaining a computational time comparable to Greedy. Although LURE has the most aggressive hallucination suppression, its computational time is 9 times that of Greedy, making it impractical for real-time inference in real-world applications. This is because LURE employs a fine-tuned 13B LVLM as revisor to correct captions generated by the 7B model. On one hand, this introduces substantial computational costs for both fine-tuning and inference; on the other hand, such excessive revision leads to a sharp decline in F1. In comparison, attention intervention methods allow dynamic balancing between hallucination rate and accuracy by adjusting the amplification factor $\alpha$. Combined with their negligible additional computational overhead, attention intervention methods demonstrate strong practical applicability.

\subsection{Hyperparameter Ablation Analysis}
Since the reduction of hallucination rates inevitably accompanies a decline in predictive performance or linguistic quality, blindly pursuing lower hallucination rates is undesirable. We randomly selected 500 images from the COCO dataset to generate captions and evaluated the hallucination rate via the $C_S$, $C_I$, prediction performance via F1 score, and language quality via $D_1$.
We evaluated PAI, HGAI, IAT, and AdaIAT on LLaVA‑1.5‑7B, LLaVA‑1.5‑13B, Janus‑Pro‑7B, and Qwen2.5‑VL‑7B. Performance metrics under varying hyperparameter settings are reported in Tab.~\ref{tab:baseline_llava7b}-\ref{tab:adaiat_qwen}, thereby quantifying each method’s trade‑offs among hallucination rate, predictive capability, and language quality across different models. To ensure a fair comparison, we report in the main text the metrics corresponding to the lowest achievable hallucination rate while maintaining F1 without significant degradation.


\subsection{Experimental Results on OpenCHAIR}
To further investigate the generalization capability of AdaIAT, this section presents full experimental results on OpenCHAIR. The CHAIR metric primarily relies on a closed vocabulary of objects (the 80 common object categories in MS-COCO), making it unable to detect open-vocabulary hallucinations (such as uncommon objects like "pearl" or "wheelchair"). Therefore, OpenCHAIR extends the CHAIR metric by relaxing its strong reliance on a closed vocabulary. It is achieved by utilizing LLaMA-2~\cite{touvron2023llama2} to rewrite MS-COCO descriptions, injecting diverse objects (e.g., "tricycle, corkscrew"), and then employing Stable Diffusion XL~\cite{podell2023sdxl} to generate image-description pairs, which are subsequently curated by humans to ensure high quality. During evaluation, objects mentioned in captions are first parsed, and then an LLM determines whether each object appears in the ground‑truth caption (labeling it as real, hallucinated, or uncertain), rather than relying on a fixed synonym dictionary. 
The hallucination rate is measured by $C_O$: 
\begin{equation}
C_O=\frac{\left|\{\text{hallucinated objects}\}\right|}{\left|\{\text{hallucinated objects and real objects}\}\right|},
\end{equation}

We selected 2,000 images from the OpenCHAIR dataset and generated captions using different LVLMs, prompted with "Please describe the image in detail." As shown in Tab.~\ref{tab:openchair}, AdaIAT achieves the lowest $C_O$ on both LLaVA-1.5-13B and Janus-Pro-7B, while maintaining text diversity comparable to the original Greedy decoding strategy. On Qwen2.5-VL, AdaIAT slightly underperforms PAI but still delivers a relatively strong performance. Given that PAI demonstrates degradation in language capabilities and weaker hallucination mitigation on other models (like LLaVA), AdaIAT achieves excellent hallucination mitigation across all evaluated models while preserving text diversity, achieving a superior trade-off.

\begin{table}[htbp]
  \centering
  \caption{Extended OpenCHAIR results on different LVLMs, with the best values (excluding Greedy) highlighted in bold.}
    \begin{tabular}{c|cc|cc|cc}
    \toprule
    Model & \multicolumn{2}{c|}{LLaVA-13B} & \multicolumn{2}{c|}{Janus-Pro} & \multicolumn{2}{c}{Qwen2.5-VL} \\
    \hline
    Method & $C_O \downarrow$    & $D_1$ $\uparrow$ & $C_O \downarrow$    & $D_1$ $\uparrow$ & $C_O \downarrow$    & $D_1$ $\uparrow$\\
    \hline
    Greedy & 0.283  & 0.62  & 0.327  & 0.63  & 0.349  & 0.64  \\
    PAI   & 0.276  & 0.58  & 0.321  & 0.63  & \textbf{0.303}  & \textbf{0.67}  \\
    HGAI  & 0.276  & 0.59  & 0.325  & 0.64  & 0.321  & 0.66  \\
    IAT   & 0.256  & \textbf{0.62}  & 0.310  & \textbf{0.65}  & 0.312  & 0.65  \\
    AdaIAT & \textbf{0.235}  & 0.61  & \textbf{0.298}  & 0.64  & 0.315  & 0.66  \\
    \bottomrule
    \end{tabular}%
  \label{tab:openchair}%
\end{table}%


\section{More Demo Cases}

\begin{figure*}[htbp]
    \centering
    \includegraphics[width=\linewidth]{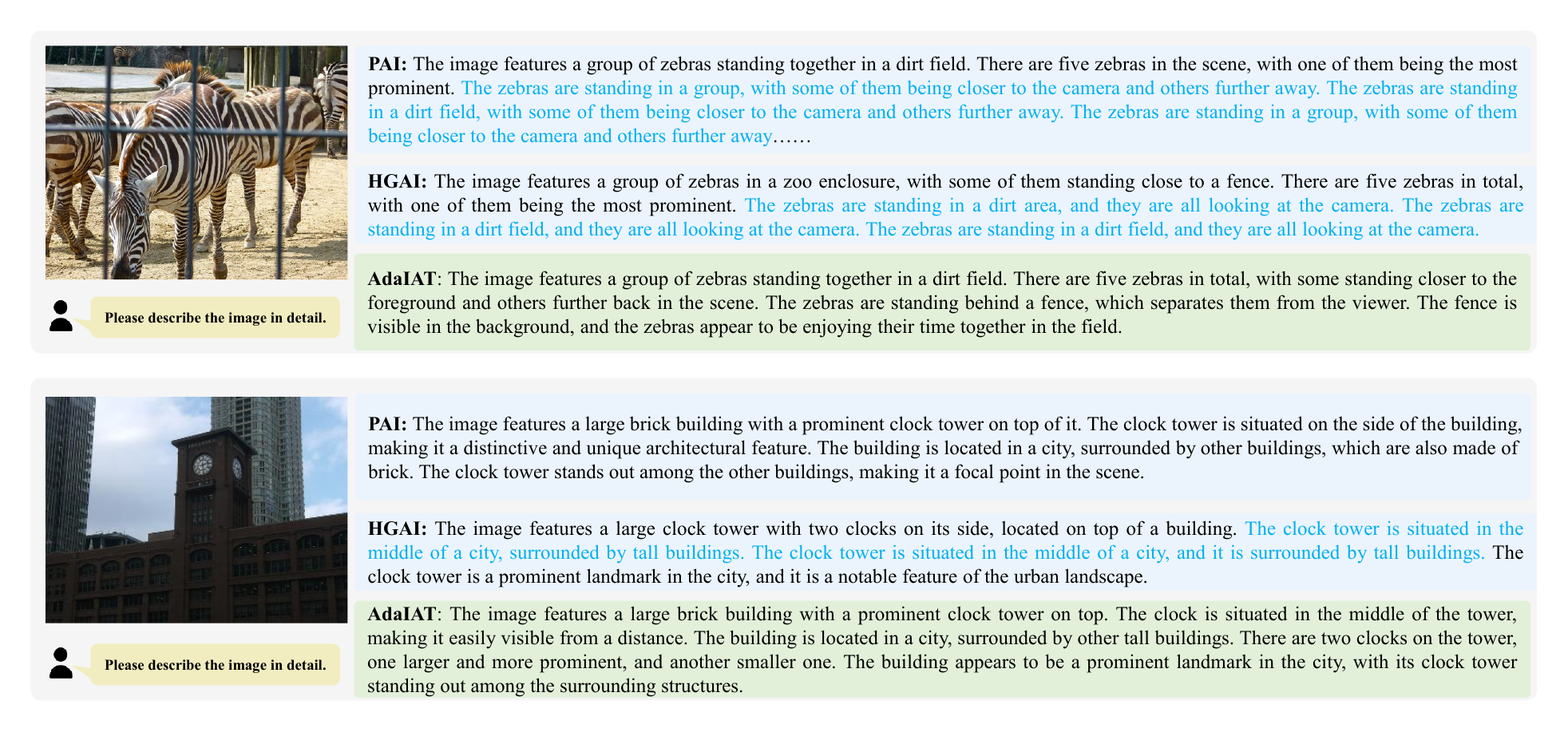}
    \caption{Demo cases of repetitive descriptions phenomenon in LLaVA-1.5-7B, with repeated segments highlighted in \textcolor{cyan}{\textbf{blue}}.}
    \label{fig:llava7b_rep}
\end{figure*}

\begin{figure*}[htbp]
    \centering
    \includegraphics[width=\linewidth]{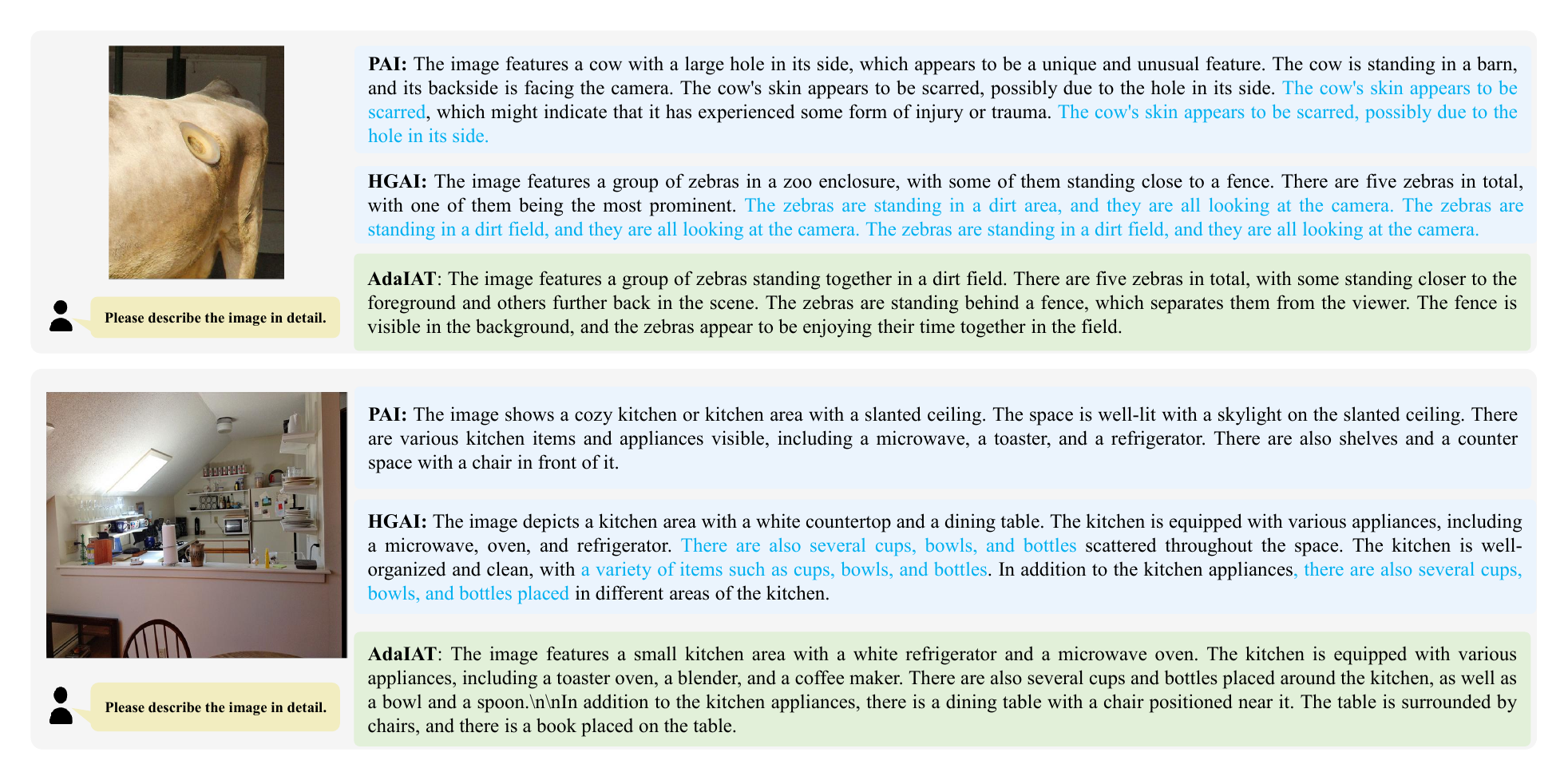}
    \caption{Demo cases of repetitive descriptions phenomenon in LLaVA-1.5-13B, with repeated segments highlighted in \textcolor{cyan}{\textbf{blue}}.}
    \label{fig:llava13b_rep}
\end{figure*}

\begin{figure*}[htbp]
    \centering
    \includegraphics[width=\linewidth]{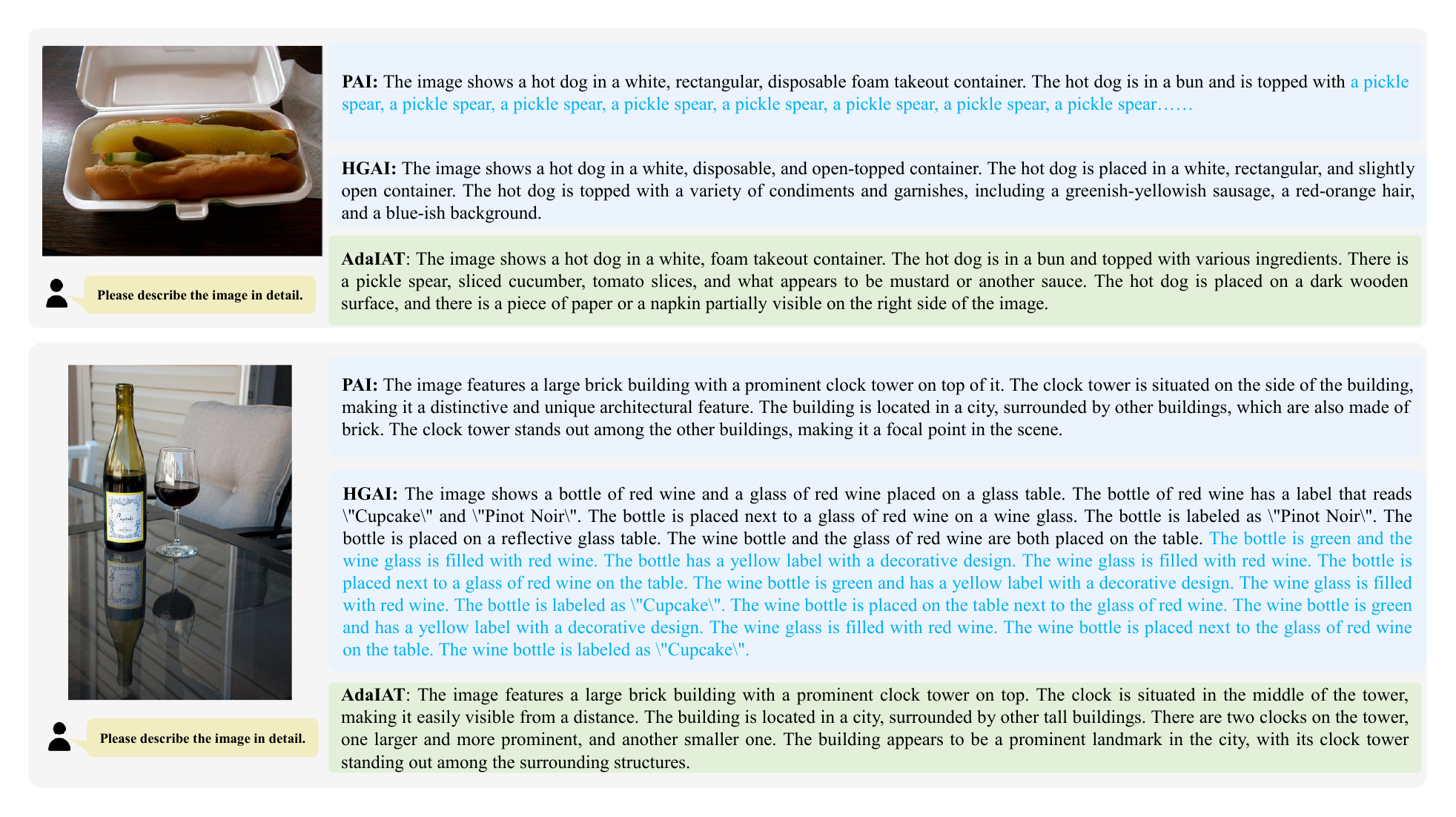}
    \caption{Demo cases of repetitive descriptions phenomenon in Janus-Pro-7B, with repeated segments highlighted in \textcolor{cyan}{\textbf{blue}}.}
    \label{fig:janus_rep}
\end{figure*}

\begin{figure*}[htbp]
    \centering
    \includegraphics[width=\linewidth]{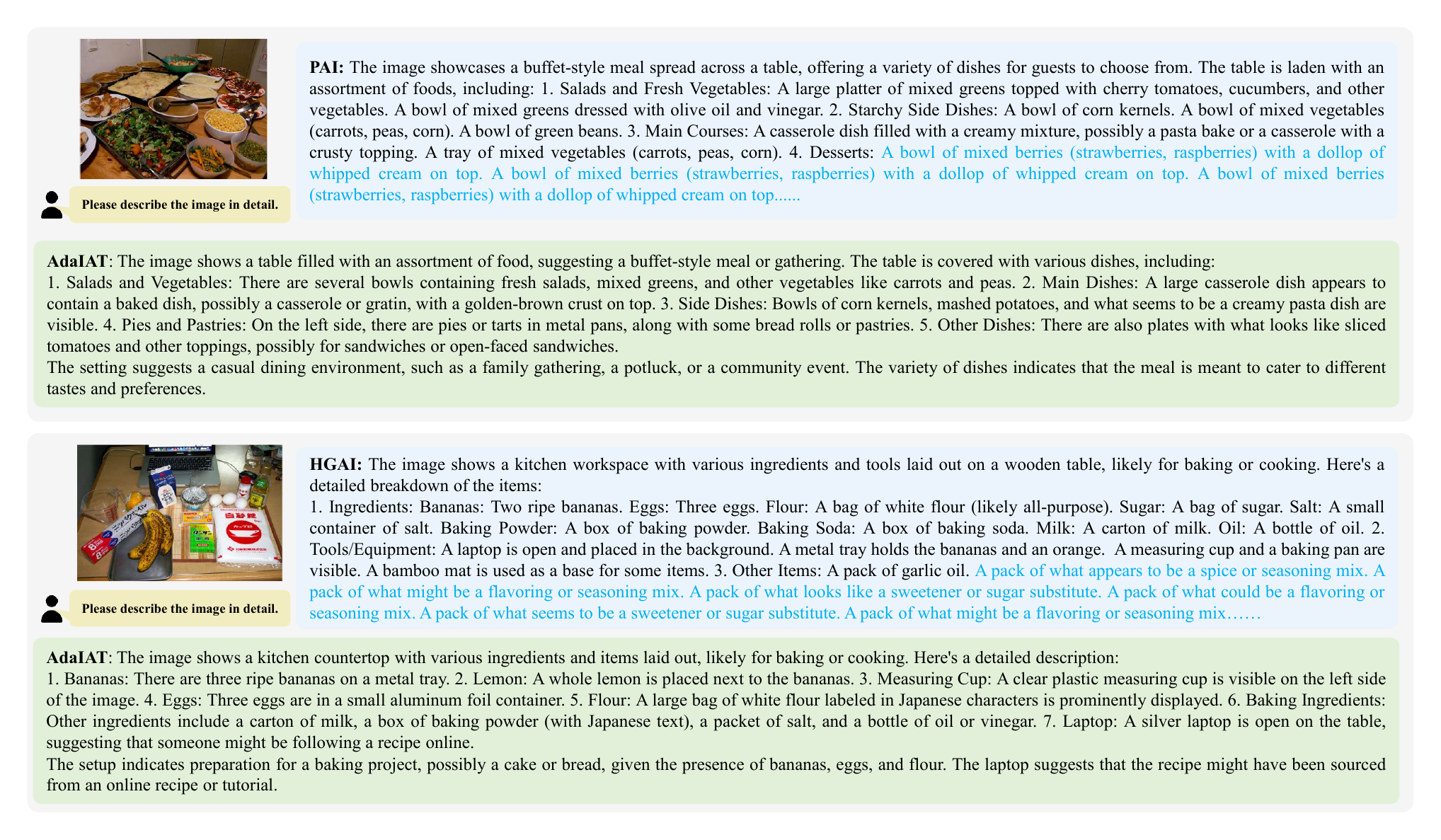}
    \caption{Demo cases of repetitive descriptions phenomenon in Qwen2.5-VL-7B, with repeated segments highlighted in \textcolor{cyan}{\textbf{blue}}.}
    \label{fig:qwen_rep}
\end{figure*}

To more intuitively demonstrate the effectiveness of AdaIAT, we present additional demo cases in this section. Fig.~\ref{fig:llava7b_rep}-\ref{fig:qwen_rep} illustrate captions generated by three methods (PAI, HGAI, and AdaIAT) across four LVLMs. 
These cases highlight the repetitive descriptions induced by existing attention intervention methods (PAI and HGAI). As shown in Fig.~\ref{fig:llava7b_rep}-\ref{fig:janus_rep}, this phenomenon often occurs when the image contains few elements or only a single salient object. 
Due to insufficient attention to generated text tokens, PAI and HGAI forget the completed descriptions of objects and persistently describe the salient object, resulting in redundant repetitions. In contrast, AdaIAT retains memory of described content, terminates descriptions promptly, and maintains concise language without repetition.
As shown in Fig.~\ref{fig:qwen_rep}, with long-text-preferring models like Qwen2.5-VL, the model struggles to incorporate extensive preceding textual information due to insufficient attention, therefore losing logical language generation capabilities. 

\begin{figure*}[htbp]
    \centering
    \includegraphics[width=\linewidth]{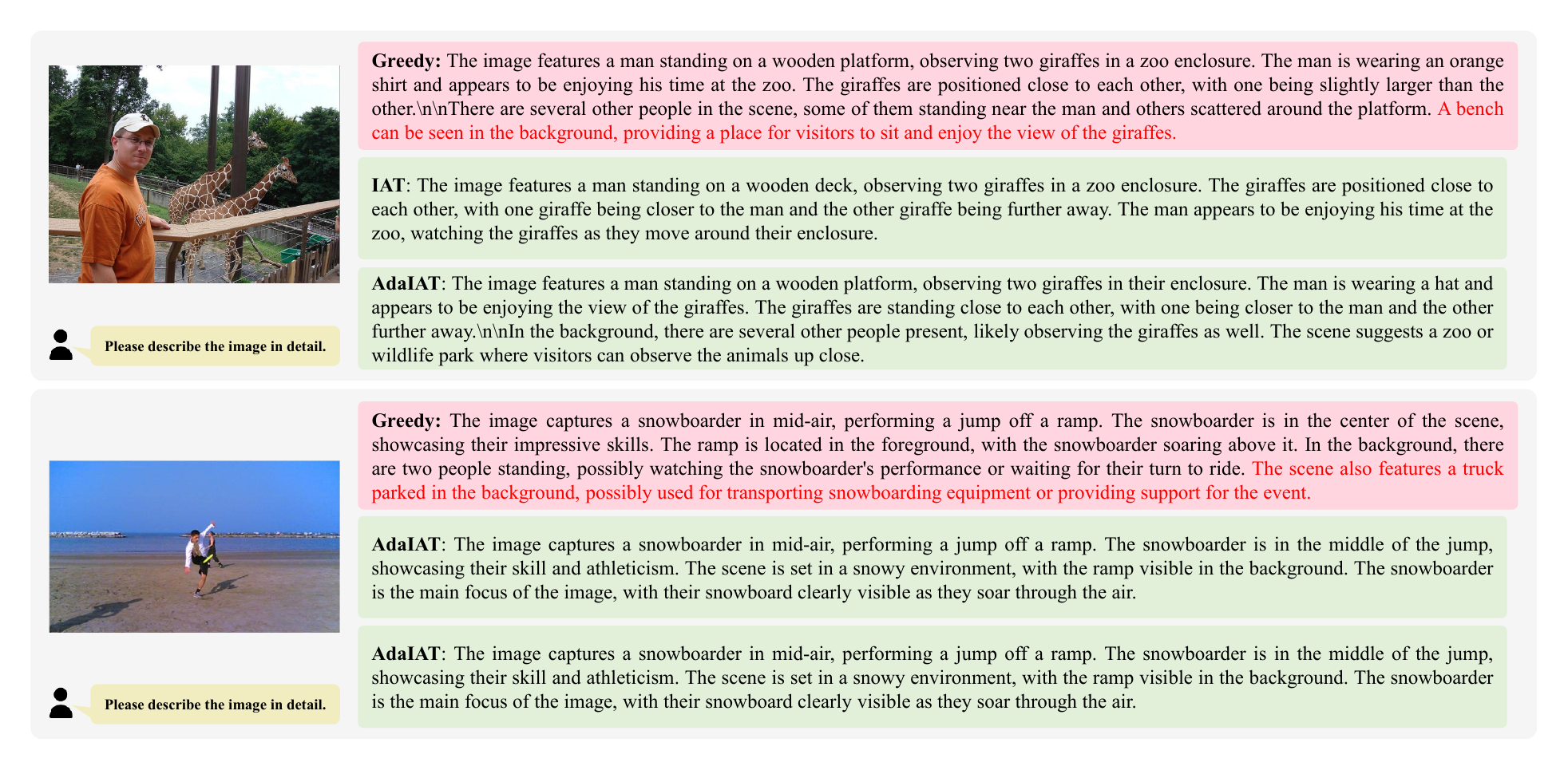}
    \caption{Qualitative results of hallucination mitigation in LLaVA-1.5-7B,, with hallucinated segments highlighted in \textcolor{red}{\textbf{red}}.}
    \label{fig:llava7b_hul}
\end{figure*}

\begin{figure*}[htbp]
    \centering
    \includegraphics[width=\linewidth]{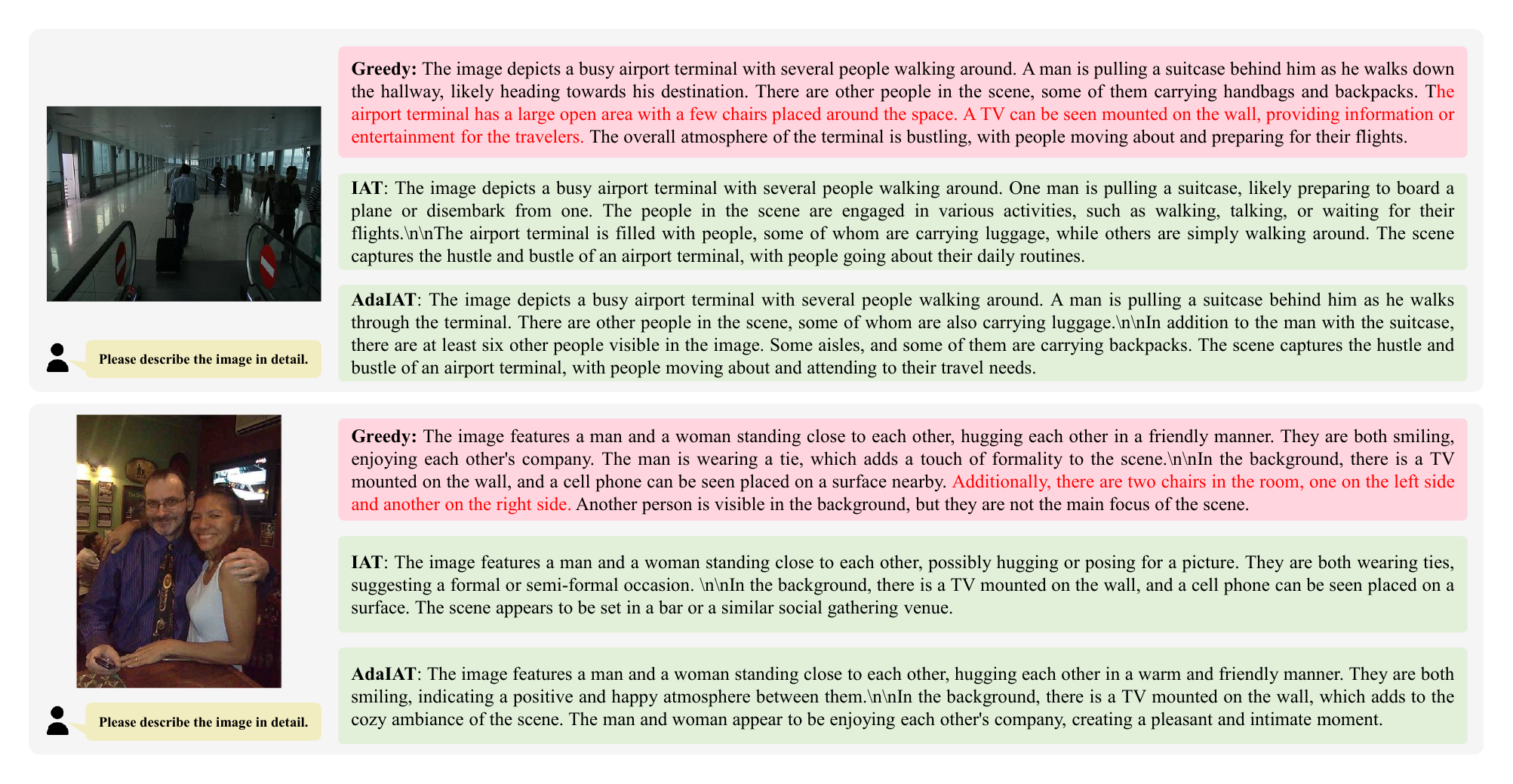}
    \caption{Qualitative results of hallucination mitigation in LLaVA-1.5-13B, with hallucinated segments highlighted in \textcolor{red}{\textbf{red}}.}
    \label{fig:llava13b_hul}
\end{figure*}

\begin{figure*}[htbp]
    \centering
    \includegraphics[width=\linewidth]{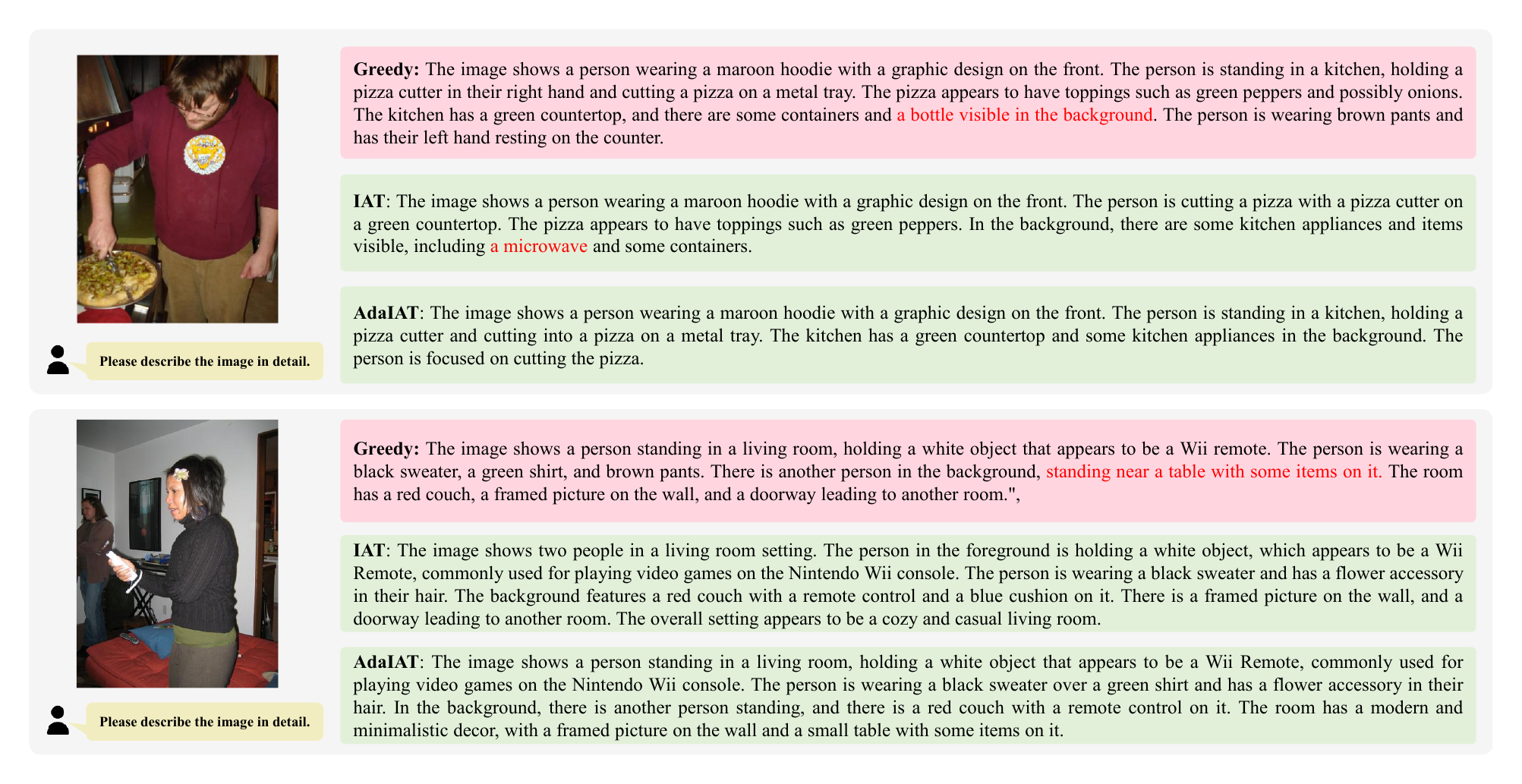}
    \caption{Qualitative results of hallucination mitigation in Janus-Pro-7B, with hallucinated segments highlighted in \textcolor{red}{\textbf{red}}.}
    \label{fig:janus_hul}
\end{figure*}

\begin{figure*}[htbp]
    \centering
    \includegraphics[width=\linewidth]{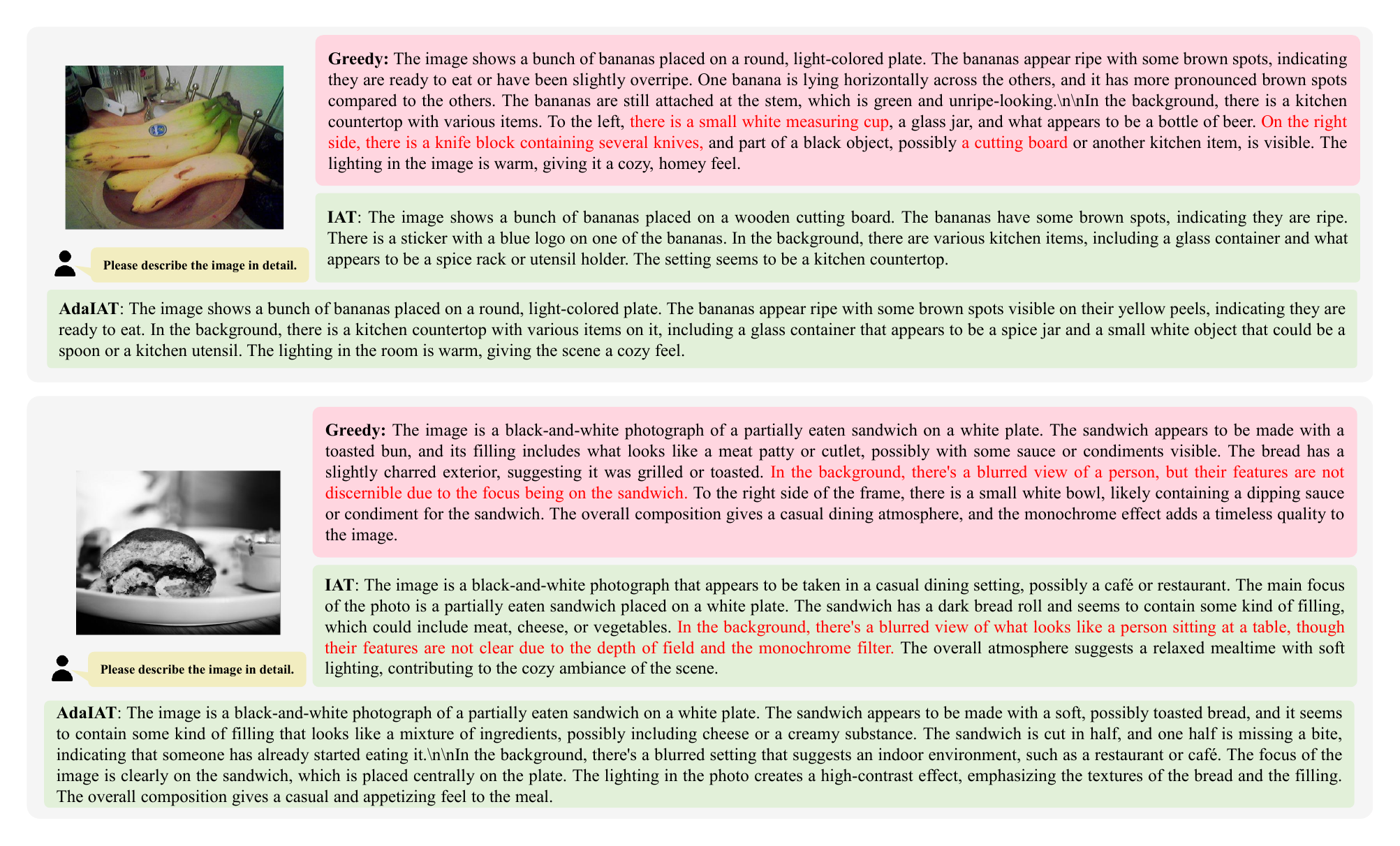}
    \caption{Qualitative results of hallucination mitigation in Qwen2.5-VL-7B, with hallucinated segments highlighted in \textcolor{red}{\textbf{red}}.}
    \label{fig:qwen_hul}
\end{figure*}

To demonstrate the hallucination mitigation effect of AdaIAT, Fig.~\ref{fig:llava7b_hul}-\ref{fig:qwen_hul} present demo cases of captions generated by Greedy, IAT, and AdaIAT methods across the above four models. Overall, IAT still occasionally exhibits hallucinations (Fig.~\ref{fig:janus_hul} and Fig.~\ref{fig:qwen_hul}), whereas AdaIAT delivers markedly superior hallucination mitigation. Even for models like Janus-Pro (Fig.~\ref{fig:janus_hul}) that exhibit extremely low hallucination levels, AdaIAT effectively eliminates sporadic, subtle hallucinations, showcasing its efficacy.


\begin{table*}[htbp]
  \centering
  \caption{Ablation analysis results for the amplification factor $\alpha$ on HGAI, PAI, and IAT in LLaVA-1.5-7B. Values highlighted in \textbf{bold} indicate those selected for reporting. The symbol "\textbackslash{}" denotes that the model lost normal language generation capabilities, rendering metric evaluation infeasible.}
    \begin{tabular}{c|cccc|cccc|cccc}
    \toprule
    
         Method & \multicolumn{4}{c|}{HGAI}     & \multicolumn{4}{c|}{PAI}      & \multicolumn{4}{c}{IAT} \\
    \hline
    $\alpha$ & $C_S \downarrow$ & $C_I\downarrow$ & F1 $\uparrow$    & $D_1\uparrow$   & $C_S\downarrow$ & $C_I\downarrow$ & F1  $\uparrow$  & $D_1\uparrow$   & $C_S\downarrow$ & $C_I\downarrow$ & F1 $\uparrow$   & $D_1\uparrow$  \\
    \hline
    0.2   & 48.4  & 12.6  & 78.7  & 0.58  & 48.2  & 13.2  & 78.1  & 0.59  & 46.4  & 12.9  & 77.9  & 0.61  \\
    0.3   & 49.0  & 12.1  & 79.1  & 0.57  & 48.0  & 12.0  & 78.5  & 0.59  & 46.6  & 12.8  & 77.6  & 0.61  \\
    0.4   & 43.8  & 11.1  & 78.9  & 0.55  & 45.0  & 11.7  & 79.2  & 0.57  & 42.6  & 12.1  & 78.2  & 0.61  \\
    0.5   & \textbf{31.4}  & \textbf{6.9}   & \textbf{78.3}  & \textbf{0.50}  & \textbf{31.8}  & \textbf{7.8}   & \textbf{77.7}  & \textbf{0.50}  & 40.6  & 10.7  & 78.1  & 0.61  \\
    0.6   & 7.2   & 4.1   & 69.3  & 0.59  & 9.4   & 3.2   & 68.5  & 0.28  & 36.4  & 10.3  & 78.1  & 0.62  \\
    0.7   & 2.8   & 1.9   & 58.0  & 0.88  & 4.4   & 3.5   & 49.9  & 0.90  & 35.2  & 10.5  & 77.4  & 0.61  \\
    0.8   & 0.4   & 0.2   & 37.6  & 0.61  & 0.8   & 3.5   & 10.2  & 0.78  & \textbf{29.8}  & \textbf{9.0}   & \textbf{76.8}  & \textbf{0.61}  \\
    0.9   & \textbackslash{} & \textbackslash{} & \textbackslash{} & \textbackslash{} & \textbackslash{} & \textbackslash{} & \textbackslash{} & \textbackslash{} & 21.2  & 7.0   & 73.5  & 0.57  \\
    \bottomrule
    \end{tabular}%
    \vspace{-10pt}
  \label{tab:baseline_llava7b}%
\end{table*}%

\begin{table*}[htbp]
  \centering
  \caption{Ablation analysis results of the balance coefficient $\beta$ and amplification factor $\alpha$ for AdaIAT on LLaVA-1.5-7B. Values highlighted in \textbf{bold} indicate those selected for reporting.}
    \begin{tabular}{c|cccc|cccc|cccc}
    \toprule
    $\beta$  & \multicolumn{4}{c|}{0.25}     & \multicolumn{4}{c|}{0.5}      & \multicolumn{4}{c}{1} \\
    \hline
    $\alpha$ & $C_S \downarrow$ & $C_I\downarrow$ & F1 $\uparrow$    & $D_1\uparrow$   & $C_S\downarrow$ & $C_I\downarrow$ & F1  $\uparrow$  & $D_1\uparrow$   & $C_S\downarrow$ & $C_I\downarrow$ & F1 $\uparrow$   & $D_1\uparrow$  \\
    \hline
    3     & 36.0  & 9.5   & 79.3  & 0.61  & 36.2  & 9.5   & 79.0  & 0.61  & 35.8  & 10.0  & 78.9  & 0.60  \\
    4     & 35.2  & 9.2   & 79.2  & 0.61  & 34.6  & 9.5   & 78.9  & 0.60  & 36.2  & 10.1  & 78.5  & 0.60  \\
    5     & 32.8  & 8.9   & 79.2  & 0.60  & 33.8  & 9.3   & 78.4  & 0.60  & 35.2  & 9.6   & 78.0  & 0.60  \\
    6     & 32.4  & 8.9   & 78.7  & 0.60  & \textbf{31.4} & \textbf{8.3} & \textbf{79.4} & \textbf{0.60 } & 32.8  & 8.9   & 78.6  & 0.60  \\
    7     & 31.8  & 9.4   & 77.7  & 0.60  & 31.8  & 8.6   & 78.7  & 0.59  & 32.6  & 8.8   & 77.8  & 0.60  \\
    8     & 31.4  & 9.1   & 77.9  & 0.59  & 29.4  & 7.9   & 79.0  & 0.59  & 30.4  & 8.4   & 78.5  & 0.59  \\
    9     & 29.4  & 8.3   & 78.5  & 0.59  & 30.0  & 7.8   & 78.3  & 0.58  & 27.4  & 8.0   & 77.9  & 0.59  \\
    \bottomrule
    \end{tabular}%
    \vspace{-10pt}
  \label{tab:adaiat_7b}%
\end{table*}%

\begin{table*}[htbp]
  \centering
  \caption{Ablation analysis results for the amplification factor $\alpha$ on HGAI, PAI, and IAT in LLaVA-1.5-13B. Values highlighted in \textbf{bold} indicate those selected for reporting. The symbol "\textbackslash{}" denotes that the model lost normal language generation capabilities, rendering metric evaluation infeasible.}
    \begin{tabular}{c|cccc|cccc|cccc}
    \toprule
          & \multicolumn{4}{c|}{HGAI}     & \multicolumn{4}{c|}{PAI}      & \multicolumn{4}{c}{IAT} \\
    \hline
    $\alpha$ & $C_S \downarrow$ & $C_I\downarrow$ & F1 $\uparrow$    & $D_1\uparrow$   & $C_S\downarrow$ & $C_I\downarrow$ & F1  $\uparrow$  & $D_1\uparrow$   & $C_S\downarrow$ & $C_I\downarrow$ & F1 $\uparrow$   & $D_1\uparrow$  \\
    \hline
    0.2   & 50.8  & 13.0  & 78.5  & 0.59  & 47.6  & 12.5  & 79.0  & 0.59  & 44.0  & 11.8  & 78.9  & 0.61  \\
    0.3   & 49.0  & 12.7  & 79.0  & 0.58  & 46.4  & 12.7  & 78.4  & 0.58  & 40.4  & 11.1  & 79.5  & 0.61  \\
    0.4   & 42.8  & 11.6  & 78.6  & 0.56  & 44.4  & 11.9  & 79.1  & 0.58  & 40.8  & 10.8  & 79.8  & 0.61  \\
    0.5   & \textbf{26.4} & \textbf{7.8}  & \textbf{76.9}  & \textbf{0.55}  & \textbf{34.2}  & \textbf{9.0}  & \textbf{79.3} & \textbf{0.56}  & 36.8  & 10.1  & 79.6  & 0.62  \\
    0.6   & 10.4  & 5.2   & 69.8  & 0.73  & 19.2  & 6.3   & 74.6  & 0.53  & 37.6  & 10.5  & 78.7  & 0.62  \\
    0.7   & 3.6   & 3.6   & 51.5  & 0.63  & 7.0   & 6.8   & 41.8  & 0.66  & 36.6  & 9.9   & 78.6  & 0.62  \\
    0.8   & 2.4   & 1.8   & 24.4  & 0.29  & \textbackslash{} & \textbackslash{} & \textbackslash{} & \textbackslash{} & \textbf{31.0}  & \textbf{8.6} & \textbf{78.6} & \textbf{0.61} \\
    0.9   & \textbackslash{} & \textbackslash{} & \textbackslash{} & \textbackslash{} & \textbackslash{} & \textbackslash{} & \textbackslash{} & \textbackslash{} & 26.2  & 7.5   & 76.7  & 0.55  \\
    \bottomrule
    \end{tabular}%
    \vspace{-10pt}
  \label{tab:addlabel}%
\end{table*}%

\begin{table*}[htbp]
  \centering
  \caption{Ablation analysis results of the balance coefficient $\beta$ and amplification factor $\alpha$ for AdaIAT on LLaVA-1.5-13B. Values highlighted in \textbf{bold} indicate those selected for reporting.}
    \begin{tabular}{c|cccc|cccc|cccc}
    \toprule
    $\beta$  & \multicolumn{4}{c|}{0.25}     & \multicolumn{4}{c|}{0.5}      & \multicolumn{4}{c}{1} \\
    \hline
    $\alpha$ & $C_S \downarrow$ & $C_I\downarrow$ & F1 $\uparrow$    & $D_1\uparrow$   & $C_S\downarrow$ & $C_I\downarrow$ & F1  $\uparrow$  & $D_1\uparrow$   & $C_S\downarrow$ & $C_I\downarrow$ & F1 $\uparrow$   & $D_1\uparrow$  \\
    \hline
    9     & 32.4  & 8.7   & 78.3  & 0.61  & 30.2  & 8.0   & 78.7  & 0.61  & 30.4  & 7.9   & 78.9  & 0.61  \\
    10    & 31.6  & 8.2   & 78.9  & 0.60  & 28.4  & 7.3   & 79.5  & 0.60  & 28.8  & 7.4   & 78.8  & 0.60  \\
    11    & 31.0  & 8.2   & 78.0  & 0.60  & 28.8  & 8.1   & 78.8  & 0.60  & 26.6  & 7.4   & 79.0  & 0.60  \\
    12    & 30.2  & 7.6   & 78.0  & 0.60  & 26.8  & 6.9   & 79.3  & 0.60  & \textbf{25.2} & \textbf{6.1} & \textbf{79.2} & \textbf{0.60} \\
    13    & 29.0  & 7.6   & 78.0  & 0.60  & 28.0  & 7.3   & 77.7  & 0.59  & 25.8  & 7.3   & 78.4  & 0.60  \\
    14    & 27.8  & 7.8   & 78.3  & 0.60  & 27.4  & 8.0   & 77.9  & 0.59  & 24.2  & 6.9   & 78.7  & 0.60  \\
    15    & 26.6  & 6.9   & 78.1  & 0.59  & 23.2  & 6.7   & 78.1  & 0.59  & 23.8  & 8.7   & 78.3  & 0.59  \\
    \bottomrule
    \end{tabular}%
    \vspace{-10pt}
  \label{tab:addlabel}%
\end{table*}%

\begin{table*}[htbp]
  \centering
  \caption{Ablation analysis results for the amplification factor $\alpha$ on HGAI, PAI, and IAT in Janus-Pro-7B. Values highlighted in \textbf{bold} indicate those selected for reporting. The symbol "\textbackslash{}" denotes that the model lost normal language generation capabilities, rendering metric evaluation infeasible.}
    \begin{tabular}{c|cccc|cccc|cccc}
    \toprule
          & \multicolumn{4}{c|}{HGAI}     & \multicolumn{4}{c|}{PAI}      & \multicolumn{4}{c}{IAT} \\
    \hline
    $\alpha$ & $C_S \downarrow$ & $C_I\downarrow$ & F1 $\uparrow$    & $D_1\uparrow$   & $C_S\downarrow$ & $C_I\downarrow$ & F1  $\uparrow$  & $D_1\uparrow$   & $C_S\downarrow$ & $C_I\downarrow$ & F1 $\uparrow$   & $D_1\uparrow$  \\
    \hline
    0.1   & 26.6  & 6.9   & 77.0  & 0.62  & 26.6  & 6.8   & 77.3  & 0.62  & 25.8  & 6.7   & 76.5  & 0.62  \\
    0.2   & 28.0  & 7.5   & 77.0  & 0.62  & 27.4  & 7.0   & 76.9  & 0.62  & 26.2  & 6.6   & 76.6  & 0.63  \\
    0.3   & 26.4  & 6.7   & 76.3  & 0.62  & 27.2  & 6.9   & 77.3  & 0.61  & 22.6  & 6.0   & 76.7  & 0.63  \\
    0.4   & \textbf{21.0}  & \textbf{5.3}   & \textbf{75.9}  & \textbf{0.62}  & \textbf{20.4}  & \textbf{5.6}   & \textbf{76.1}  & \textbf{0.61}  & 22.8  & 6.0   & 76.3  & 0.63  \\
    0.5   & 18.0  & 6.0   & 68.0  & 0.60  & 11.6  & 4.4   & 67.7  & 0.61  & 21.2  & 5.7   & 76.1  & 0.64  \\
    0.6   & 11.8  & 3.4   & 57.4  & 0.43  & 1.2   & 1.4   & 26.5  & 0.62  & \textbf{20.6}  & \textbf{5.3}   & \textbf{75.3}  & \textbf{0.65}  \\
    0.7   & 8.8   & 3.1   & 51.6  & 0.23  & \textbackslash{} & \textbackslash{} & \textbackslash{} & \textbackslash{} & 17.4  & 5.1   & 73.8  & 0.65  \\
    0.8   & 6.6   & 6.1   & 34.5  & 0.18  & \textbackslash{} & \textbackslash{} & \textbackslash{} & \textbackslash{} & 13.2  & 8.9   & 69.0  & 0.64  \\
    \bottomrule
    \end{tabular}%
    \vspace{-10pt}
  \label{tab:addlabel}%
\end{table*}%

\begin{table*}[htbp]
  \centering
  \caption{Ablation analysis results of the balance coefficient $\beta$ and amplification factor $\alpha$ for AdaIAT on Janus-Pro-7B. Values highlighted in \textbf{bold} indicate those selected for reporting.}
    \begin{tabular}{c|cccc|cccc|cccc}
    \toprule
    $\beta$  & \multicolumn{4}{c|}{0.25}     & \multicolumn{4}{c|}{0.5}      & \multicolumn{4}{c}{1} \\
    \hline
    $\alpha$ & $C_S \downarrow$ & $C_I\downarrow$ & F1 $\uparrow$    & $D_1\uparrow$   & $C_S\downarrow$ & $C_I\downarrow$ & F1  $\uparrow$  & $D_1\uparrow$   & $C_S\downarrow$ & $C_I\downarrow$ & F1 $\uparrow$   & $D_1\uparrow$  \\
    \hline
    1     & 24.8  & 6.4   & 76.3  & 0.63  & 24.8  & 6.2   & 75.8  & 0.64  & 25.2  & 6.5   & 76.2  & 0.63  \\
    2     & 21.8  & 5.9   & 75.7  & 0.64  & 22.8  & 6.1   & 75.3  & 0.64  & 22.0  & 6.1   & 76.3  & 0.64  \\
    3     & 20.0  & 6.0   & 76.2  & 0.64  & 21.4  & 6.0   & 75.4  & 0.64  & 21.6  & 5.8   & 75.9  & 0.64  \\
    4     & 17.6  & 4.9   & 75.7  & 0.64  & 20.8  & 5.7   & 75.1  & 0.64  & \textbf{19.0}  & \textbf{4.9}   & \textbf{76.5}  & \textbf{0.64}  \\
    5     & 19.4  & 5.1   & 75.2  & 0.65  & 20.4  & 5.5   & 75.0  & 0.64  & 17.8  & 4.6   & 75.4  & 0.65  \\
    6     & 16.2  & 4.4   & 74.9  & 0.64  & 18.0  & 4.7   & 75.1  & 0.65  & 18.0  & 5.7   & 74.2  & 0.65  \\
    7     & 16.6  & 4.8   & 73.8  & 0.65  & 19.0  & 5.1   & 74.3  & 0.65  & 18.2  & 5.6   & 74.1  & 0.64  \\
    \bottomrule
    \end{tabular}%
    \vspace{-10pt}
  \label{tab:addlabel}%
\end{table*}%

\begin{table*}[htbp]
  \centering
  \caption{Ablation analysis results for the amplification factor $\alpha$ on HGAI, PAI, and IAT in Qwen2.5-VL-7B. Values highlighted in \textbf{bold} indicate those selected for reporting.}
    \begin{tabular}{c|cccc|cccc|cccc}
    \toprule
          & \multicolumn{4}{c|}{HGAI}     & \multicolumn{4}{c|}{PAI}      & \multicolumn{4}{c}{IAT} \\
    \hline
    $\alpha$ & $C_S \downarrow$ & $C_I\downarrow$ & F1 $\uparrow$    & $D_1\uparrow$   & $C_S\downarrow$ & $C_I\downarrow$ & F1  $\uparrow$  & $D_1\uparrow$   & $C_S\downarrow$ & $C_I\downarrow$ & F1 $\uparrow$   & $D_1\uparrow$  \\
    \hline
    0.1   & 36.6  & 8.5   & 77.0  & 0.65  & 37.0  & 9.0   & 75.7  & 0.65  & 32.0  & 8.0   & 75.9  & 0.66  \\
    0.2   & 34.4  & 8.1   & 76.9  & 0.65  & 38.4  & 10.2  & 74.9  & 0.65  & \textbf{29.2}  & \textbf{7.7}   & \textbf{76.4}  & \textbf{0.67}  \\
    0.3   & 33.8  & 8.2   & 77.3  & 0.64  & 37.0  & 9.2   & 74.6  & 0.64  & 23.6  & 6.9   & 76.0  & 0.67  \\
    0.4   & \textbf{30.4}  & \textbf{8.0}   & \textbf{76.8}  & \textbf{0.65}  & 37.2  & 9.0   & 73.6  & 0.63  & 25.2  & 6.9   & 75.8  & 0.69  \\
    0.5   & 30.4  & 7.4   & 76.0  & 0.65  & \textbf{32.0}  & \textbf{8.2}  & \textbf{74.7}  & \textbf{0.64}  & 24.0  & 7.0   & 74.1  & 0.70  \\
    0.6   & 27.0  & 8.2   & 73.9  & 0.66  & 24.8  & 6.9   & 69.9  & 0.65  & 20.8  & 6.0   & 72.6  & 0.70  \\
    0.7   & 16.4  & 6.2   & 67.9  & 0.76  & 9.4   & 4.5   & 58.6  & 0.79  & 18.0  & 6.0   & 71.4  & 0.71  \\
    0.8   & 6.6   & 4.1   & 57.4  & 0.85  & 3.6   & 2.2   & 52.5  & 0.87  & 18.6  & 5.8   & 70.5  & 0.70  \\
    \bottomrule
    \end{tabular}%
    \vspace{-10pt}
  \label{tab:baseline_qwen}%
\end{table*}%

\begin{table*}[htbp]
  \centering
  \caption{Ablation analysis results of the balance coefficient $\beta$ and amplification factor $\alpha$ for AdaIAT on Qwen2.5-VL. Values highlighted in \textbf{bold} indicate those selected for reporting.}
    \begin{tabular}{c|cccc|cccc|cccc}
    \toprule
    $\beta$  & \multicolumn{4}{c|}{0.25}     & \multicolumn{4}{c|}{0.5}      & \multicolumn{4}{c}{1} \\
    \hline
    $\alpha$ & $C_S \downarrow$ & $C_I\downarrow$ & F1 $\uparrow$    & $D_1\uparrow$   & $C_S\downarrow$ & $C_I\downarrow$ & F1  $\uparrow$  & $D_1\uparrow$   & $C_S\downarrow$ & $C_I\downarrow$ & F1 $\uparrow$   & $D_1\uparrow$  \\
    \midrule
    6     & 31.2  & 7.3   & 76.7  & 0.66  & 30.6  & 7.3   & 77.0  & 0.66  & 31.2  & 7.6   & 76.8  & 0.66  \\
    7     & 31.2  & 7.4   & 76.2  & 0.66  & 32.8  & 7.8   & 76.6  & 0.66  & 30.8  & 7.7   & 76.8  & 0.66  \\
    8     & 31.6  & 7.6   & 76.9  & 0.66  & 30.2  & 7.5   & 76.8  & 0.66  & 31.4  & 7.4   & 77.1  & 0.66  \\
    9     & \textbf{28.4} & \textbf{6.9} & \textbf{76.8} & \textbf{0.66} & 31.2  & 7.7   & 76.7  & 0.66  & 30.2  & 7.4   & 76.7  & 0.66  \\
    10    & 30.6  & 7.6   & 76.2  & 0.66  & 33.6  & 8.4   & 75.9  & 0.66  & 29.4  & 7.6   & 76.9  & 0.66  \\
    11    & 29.8  & 7.2   & 76.9  & 0.66  & 29.6  & 7.1   & 76.9  & 0.66  & 30.6  & 7.6   & 77.1  & 0.66  \\
    12    & 30.4  & 7.4   & 76.3  & 0.66  & 30.6  & 7.8   & 76.3  & 0.66  & 30.2  & 7.4   & 77.2  & 0.66  \\
    \bottomrule
    \end{tabular}%
    \vspace{-10pt}
  \label{tab:adaiat_qwen}%
\end{table*}%

\clearpage
{
    \small
    \bibliographystyle{ieeenat_fullname}
    \bibliography{main}
}


%% file: sec/0_abstract.tex
\begin{abstract}
Hallucination has been a significant impediment to the development and application of current Large Vision-Language Models (LVLMs). 
To mitigate hallucinations, one intuitive and effective way is to directly increase attention weights to image tokens during inference. Although this effectively reduces the hallucination rate, it often induces repetitive descriptions.
To address this, we first conduct an analysis of attention patterns and reveal that real object tokens tend to assign higher attention to the generated text than hallucinated ones. 
This inspires us to leverage the generated text, which contains instruction-related visual information and contextual knowledge, to alleviate hallucinations while maintaining linguistic coherence. We therefore propose \textbf{I}ncrease \textbf{A}ttention to Generated \textbf{T}ext (IAT) and demonstrate that it significantly reduces the hallucination rate while avoiding repetitive descriptions.
To prevent naive amplification from impairing the inherent prediction capabilities of LVLMs, we further explore \textit{\textbf{Ada}ptive IAT} (AdaIAT) that employs a layer-wise threshold to control intervention time and fine-grained amplification magnitude tailored to the characteristics of each attention head.
Both analysis and experiments demonstrate the effectiveness of AdaIAT. Results of several LVLMs show that AdaIAT effectively alleviates hallucination (reducing hallucination rates $C_S$ and $C_I$ on LLaVA-1.5 by 35.8\% and 37.1\%, respectively) while preserving linguistic performance and prediction capability, achieving an attractive trade-off. Code is available at \href{https://github.com/XianguiKang/AdaIAT.git}{https://github.com/XianguiKang/AdaIAT.git}.
\end{abstract}

%% file: sec/1_intro.tex
\section{Introduction}
\label{sec:intro}



Large Vision-Language Models (LVLMs) \cite{zhu2023minigpt, Qwen-VL, chen2023shikra} integrate pre-trained visual encoders with Large Language Models (LLMs) to achieve cross-modal alignment and joint representation learning, enabling applications encompassing cross-modal content generation, image captioning, embodied intelligence interaction, and similar scenarios. However, current LVLMs all face the significant challenge of hallucinations, manifested as generating descriptions inconsistent with the input visual content, which severely constrains their reliability and practical utility.

Recent studies indicate that hallucinations arise from insufficient alignment between visual and linguistic features during the cross-modal fusion process, leading LVLMs to ``ignore'' the image content~\cite{bai2024survey, tong2024eyes, devil, 3-1}. Therefore attention intervention methods (such as PAI\cite{liu2024pai}, HGAI\cite{devil}) were proposed to directly amplify the attention weights assigned to image tokens during the generation process, thereby emphasizing the importance of visual information and reducing the hallucination rate. Although such methods are intuitive and attractive, we observed that the reduced hallucination rates achieved by these methods often come at the cost of degraded linguistic capability.

\begin{figure}[t]
    \vspace{-15pt}
    \centering
    \includegraphics[width=\linewidth]{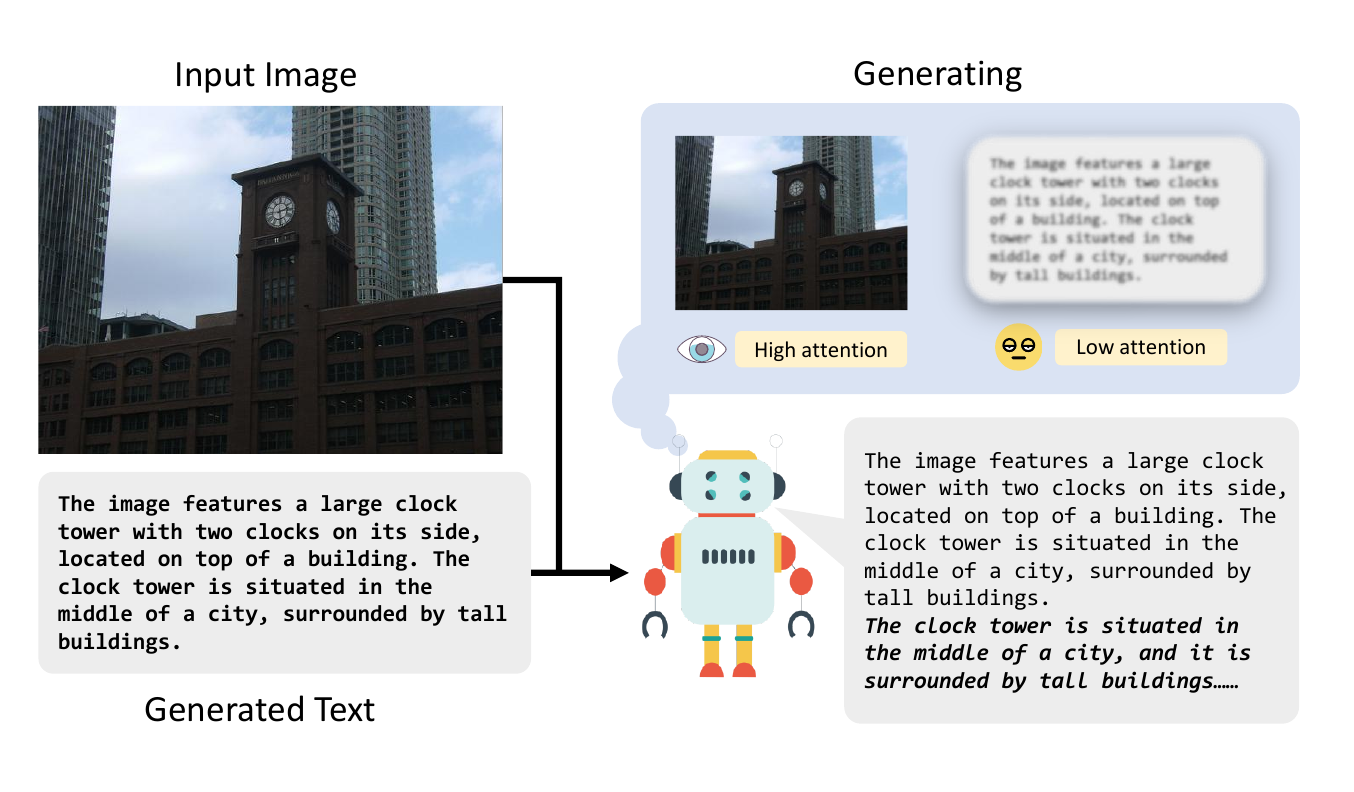}
    \vspace{-25pt}
    \caption{Case illustration: The attention intervention methods amplify the attention to image tokens, thereby emphasizing visual information and mitigating hallucinations. However, the relatively low attention to the generated text causes the model to forget preceding utterances, leading to repetitive descriptions of the prominent object `clock tower'.}
    \label{fig:intro}
    \vspace{-15pt}
\end{figure}

Fig.~\ref{fig:intro} illustrates a case of degradation in linguistic capability: PAI excessively enhances attention to the image, relatively suppressing attention to the text. This leads LVLMs to ``forget'' previously generated content, sometimes even producing output nearly identical to the preceding sentence. While such attention intervention suppresses hallucinations, it sacrifices the core linguistic capability of the LLM, deviating from the user's original intent. 
To address this issue, we first analyze the attention patterns of real versus hallucinated object sets and note that, compared to hallucinated objects,  real objects exhibit higher attention to generated text tokens ($T_p$). This insight suggests that the knowledge contained in $T_p$ may support accurate predictions. 
Intuitively, this makes sense: due to the causal attention mechanism during generation, $T_p$, as the output of the LVLM, naturally contains instruction-related visual information. As prior research~\cite{liu2024pai, devil} has shown that amplifying attention to visual information can mitigate hallucinations, would increasing attention to $T_p$ be similarly effective?



Motivated by the above observation, we propose \textit{\textbf{I}ncreased \textbf{A}ttention to Generated \textbf{T}ext} (IAT) to leverage instruction-related visual information and contextual knowledge embedded in $T_p$: the former supports more accurate image descriptions, while the latter maintains linguistic coherence and diversity. Experimental results demonstrate that IAT effectively suppresses hallucinations without introducing repetitive descriptions.
Furthermore, we notice that existing attention amplification methods typically intervene throughout the entire process regardless of whether a hallucination is present, and apply a uniform amplification magnitude that overlooks the distinct roles of different attention heads. Such coarse and unnatural intervention may disrupt the model's inherent reasoning process. To address this, we further propose \textit{\textbf{Ada}ptive IAT} (AdaIAT), which incorporates layer-wise thresholds to control the intervention timing and adaptively assigns distinct amplification magnitudes to different attention heads.

In summary, the contributions of this paper are threefold: 
1) We propose IAT, which increases attention to generated text tokens to leverage the model's prior compressed, instruction‑relevant visual representations, thereby effectively reducing hallucinations and avoiding repetitive description.
2) We further propose AdaIAT to adaptively determine both the intervention timing and amplification magnitudes, which minimizes the disruption to the inherent prediction patterns of the underlying LLM.
3) Both analysis and experiments demonstrate the effectiveness of the proposed AdaIAT. Compared with current attention intervention methods on different LVLMs, our AdaIAT significantly reduces hallucination rates while maintaining prediction capability and textual diversity.

%% file: sec/2_relatedWork.tex
\section{Related Work}
\label{Related Work}

\subsection{Large Vision-Language Models}
Advances in pre-training~\cite{pretraning} and instruction fine-tuning~\cite{finetuned} techniques have propelled the development of large language models (LLMs) such as LLaMA~\cite{llama} and Vicuna~\cite{peng2023vicuna}, which have promoted the rise of large visual language models (LVLMs). Pioneering works such as Flamingo~\cite{alayrac2022flamingo} and BLIP-2~\cite{li2023blip2} successfully adapted LLMs to visual tasks, demonstrating generative and contextual learning capabilities. Recently, visual instruction fine-tuning has further improved the capabilities of LVLMs, which usually includes a visual encoder (such as CLIP~\cite{clip}), a modality connector (aligning image features to the text domain), and LLMs.
LLaVA-1.5~\cite{llava1.5} utilizes an MLP as its linear projection layer and incorporates an AnyRes strategy to accommodate input images of varying sizes. 
Janus-Pro~\cite{chen2025janus} decouples multimodal understanding and generation via separate visual encoders and task-specific adapters, projecting their outputs into a unified sequence processed by LLM.
Qwen2.5‑VL~\cite{bai2025qwen} introduces dynamic resolution processing and absolute time encoding to accommodate complex inputs, and further integrates window attention into its vision encoder to reduce computational overhead.
Despite their remarkable capabilities, existing LVLMs generally suffer from the serious problem of hallucination, which has become an increasingly prominent research focus in recent years.

\subsection{Hallucination and Its Mitigation in LVLMs}
Hallucination in LVLMs specifically refers to the phenomenon where the text descriptions generated by LVLMs contradict the content of the input images, such as describing objects that do not exist in the image, which severely compromises model reliability. To mitigate this issue, researchers~\cite{1-1, 1-2, 1-3} have proposed multiple technical approaches: (1) Introducing additional datasets for training: compensating for data bias and cross-modal knowledge gaps by incorporating higher-quality annotated data~\cite{gunjal2024HQdata}, or enhancing vision-language alignment training~\cite{sun2023aligning}. (2) Post-hoc correction: LURE~\cite{lure} trains an LVLMs as a revisor to rewrite sentences, while Woodpecker~\cite{woodpecker} extracts key concepts for subsequent verification and correction using expert models. 
(3) Decoding strategies~\cite{2-1, 1-4}: 
VCD~\cite{leng2024vcd} mitigates over-reliance on statistical biases and language priors by contrasting output distributions generated from original versus distorted visual inputs,
whereas AGLA~\cite{an2025agla} decodes the outputs using an assembly of global and local attention. 
(4) Attention intervention: PAI~\cite{liu2024pai} globally enhances the attention on image tokens to emphasize visual information importance, while HGAI~\cite{devil} amplifies the attention to image tokens by integrating information from multiple attention heads. Unlike prior methods that often introduce substantial computational overhead during training or inference, attention intervention significantly reduces hallucination rates while maintaining lower inference costs, achieving the best performance in hallucination mitigation.
However, they may exhibit compromised language capabilities in some cases.

%% file: sec/3_method.tex
\begin{figure*}[ht]
\vspace{-10pt}
    \centering
    \includegraphics[width=0.95\linewidth]{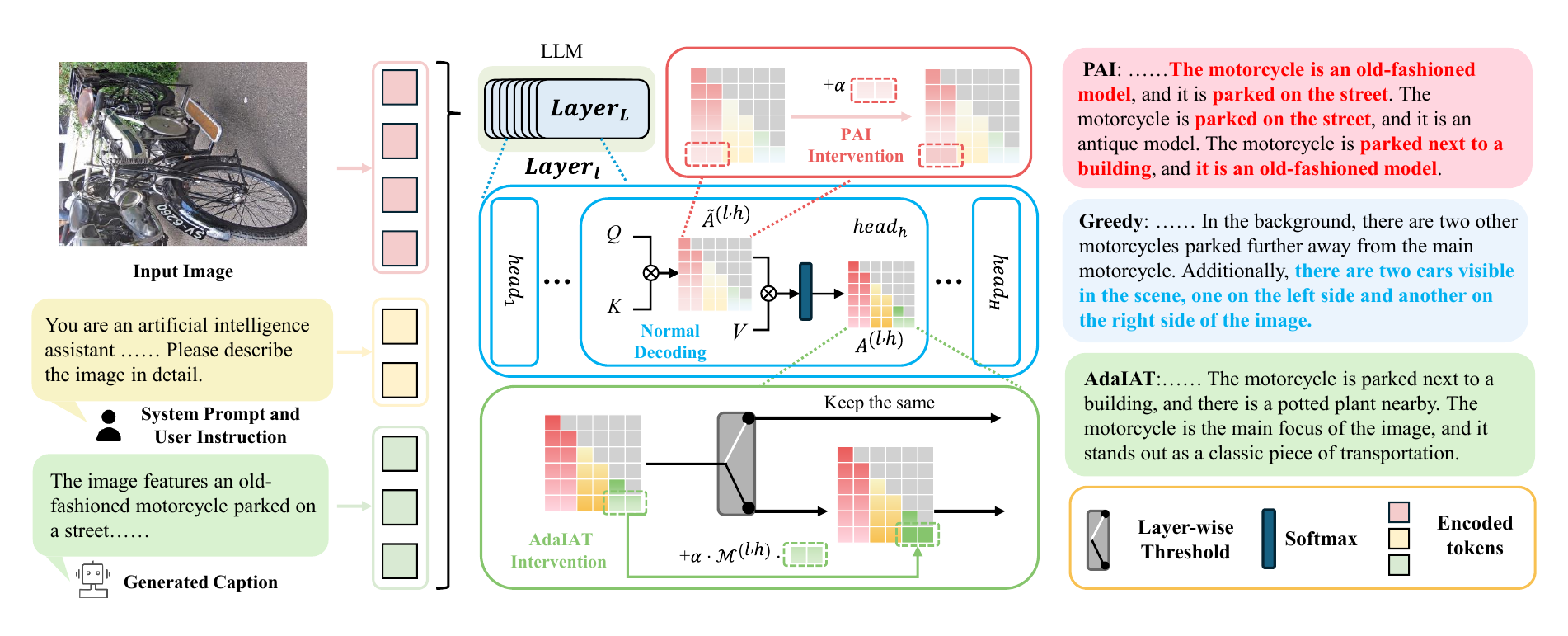}
    \vspace{-12pt}
    \caption{The mechanisms of mitigating hallucination based on Greedy in PAI and AdaIAT, with repeated descriptions annotated in \textcolor{red}{\textbf{red}} and hallucinations annotated in \textcolor{cyan}{\textbf{blue}}. While Greedy generates the hallucinated object “cars”, PAI enhances attention to image tokens with a fixed $\alpha$ to mitigate hallucination. However, it suffers from repeated subjects and monotonous, redundant language. In contrast, AdaIAT employs layer-wise thresholds to control the amplification and designs $\mathcal{M}^{(l,h)}$ for each attention head to adaptively enhance attention towards the generated text tokens, which produces accurate and hallucination-free captions.}
    \label{fig:framework}
        \vspace{-10pt}
\end{figure*}

\section{Methodology}
\label{sec:method}


The input to LVLMs comprises both text and images. LVLMs project contents from different modalities into tokens within a unified input feature space $I_{s}$ using corresponding domain-specific encoders. These tokens are then fed into the pre-trained LLM. As shown in Fig.~\ref{fig:framework}, at timestep $n$, to predict the next text token $t_{n+1}$, the LLM's input $\mathcal{I}$ consists of: system prompt tokens $S=\{s_1, \ldots, s_{a}\}$, image features tokens $V=\{v_1, \ldots, v_m\}$, user instruction tokens $U=\{u_1, \ldots, u_b\}$, and generated text tokens $T_{p}=\{t_1, \ldots, t_n\}$. The total input length is $len = a + m + b + n$. Typically, the LLM incorporates $L$ layers of attention blocks, with each layer containing $H$ attention heads. For the $h$-th head in the $l$-th layer, we focus on the self-attention computation pertaining to the token $t_n$ as follows:
\begin{equation}
\quad\boldsymbol{A}^{(l,h)}=\mathrm{softmax}(\boldsymbol{\tilde{A}}^{(l,h)})
\end{equation}
\begin{equation}
\boldsymbol{\tilde{A}}^{(l,h)}=\frac{\boldsymbol{Q}^{(l,h)}_{t_n}{(\boldsymbol{K}^{(l,h)})}^{\top}}{\sqrt{d_{k}}} 
\end{equation}
Here, $l \in (1,L)$, $h\in(1,H)$, $\boldsymbol{Q}^{(l,h)}_{t_n}$ denotes the query vector corresponding to $t_n$, and $\boldsymbol{K}^{(l,h)}$ denotes the key vectors for all tokens. During the prediction process of $t_{n+1}$, we can obtain the attention maps $\mathbf{A}\in\mathbb{R}^{L\times H \times len}$ from all layers and heads. Let $\mathbf{A}^{(l,h)}(i)$ denote the attention weight to the $i$-th token during prediction, where $i\in(1,len)$.

\begin{figure*}[t]
\vspace{-10pt}
\centering
\subfloat[$\bar{\mathbf{A}}_{T_{p}}^{r}$ and $\bar{\mathbf{A}}_{T_{p}}^{h}$ (Layers 5-18)]{
    \centering
    \includegraphics[width=0.45\linewidth]{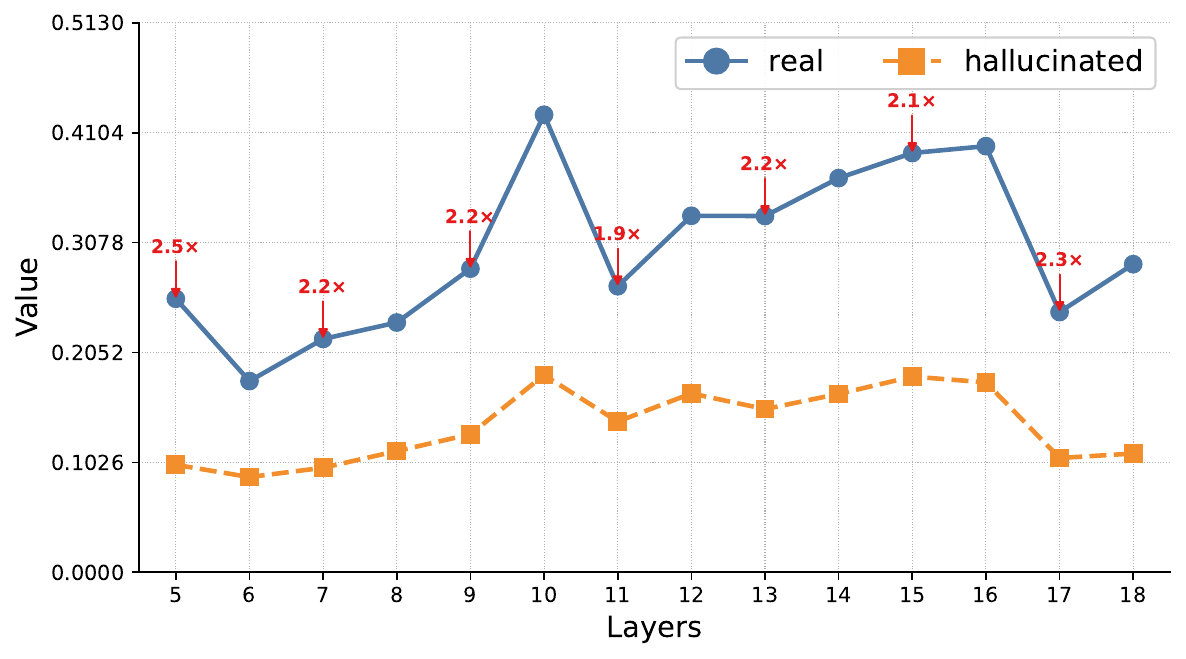}
    \label{fig:contrast_text}
}
\subfloat[$\bar{\mathbf{A}}_{V}^{r}$ and $\bar{\mathbf{A}}_{V}^{h}$ (Layers 5-18)]{
    \centering
    \includegraphics[width=0.45\linewidth]{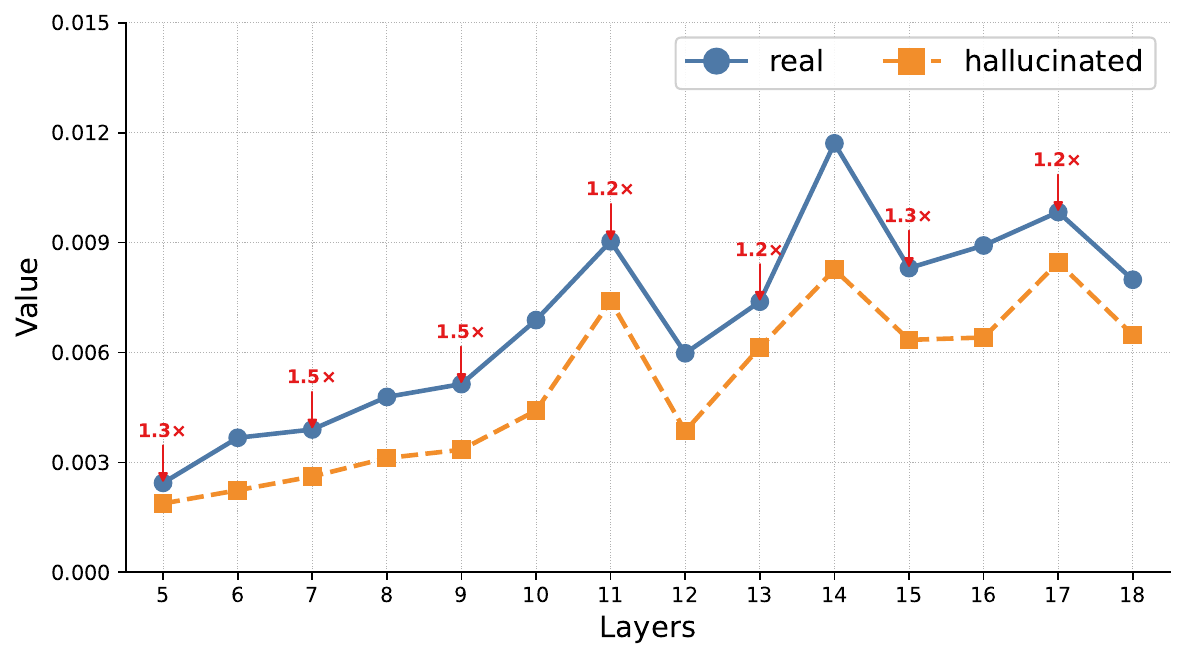}
    \label{fig:contrast_img}
}
\vspace{-8pt}
\caption{Visualization of the average per-token attention weights from text token $t_{n+1}$ to generated text tokens  $T_{p}$ ($\bar{\mathbf{A}}_{T_{p}}^{r}$ and $\bar{\mathbf{A}}_{T_{p}}^{h}$) and to image tokens $V$ ($\bar{\mathbf{A}}_{V}^{r}$ and $\bar{\mathbf{A}}_{V}^{h}$), showing only layers 5–18 for clearer observation.}
\label{fig:contrast}
\vspace{-10pt}
\end{figure*}

\subsection{Observation and Analysis}
\label{motivation}
To investigate the causes of hallucination, we analyzed the differences in internal patterns during the generation of real versus hallucinated objects. Specifically, we randomly selected 10,000 images from COCO 2014~\cite{coco} and generated captions for them using LLaVA-1.5-7B, with the prompt: ``Please describe the image in detail.'' (Results of other LVLMs are provided in the Appendix.) By identifying objects in the captions through ground truth annotations, we obtained 22,015 real objects tokens and 9,473 hallucinated objects tokens. We collected their corresponding attention weight maps subsets $D^{r}$ and $D^{h}$, and computed their average attention maps $\mathbf{A}^{r}, \mathbf{A}^{h} \in \mathbb{R}^{L \times H \times len}$. To quantify the importance of different inputs to model outputs, we aggregate attention across specific regions, for example, for $T_{p}$:
\begin{equation}
\mathbf{A}^{(l,h)}_{T_{p}}\triangleq \frac{1}{n} \sum_{i}\mathbf{A}^{(l,h)}(i), 
\quad
\mathcal{I}(i) \in T_{p}
\end{equation}
Then we obtain $\mathbf{A}_{T_{p}}\in\mathbb{R}^{L\times H}$, representing the average per-token attention weights from $t_{n+1}$ to $T_{p}$ across different layers and attention heads. Further, we aggregate attention across the heads:
\begin{equation}
\bar{\mathbf{A}}^{(l)}_{T_{p}}\triangleq \sum_{h=1}^{H}\mathbf{A}^{(l,h)}_{T_{p}}
\end{equation}
We use $ (\bar{\cdot})$ to denote the average of the heads, then we obtain $\bar{\mathbf{A}}_{T_{p}}\in\mathbb{R}^{L}$, indicating the importance of $T_{p}$ across different layers. Then, we correspondingly compute $\mathbf{A}_{V}, \bar{\mathbf{A}}_{V}$  for input image. Subsequently, we compute $\bar{\mathbf{A}}_{T_{p}}^{r}, \bar{\mathbf{A}}_{T_{p}}^{h}, \bar{\mathbf{A}}_{V}^{r}$, and $\bar{\mathbf{A}}_{V}^{h}$ from $D^{r}$ and $D^{h}$, and visualize them in Figure \ref{fig:contrast}.

Consistent with existing research~\cite{devil}, hallucination correlates with insufficient attention to $V$, manifested as $\bar{\mathbf{A}}_{V}^{r}$ consistently exceeds $\bar{\mathbf{A}}_{V}^{h}$ (typically by a factor of ×1–×1.5) in Fig.~\ref{fig:contrast_img}. 
However, we observe an intriguing phenomenon: as depicted in Fig.~\ref{fig:contrast_text}, \textbf{\textit{the discrepancy between real and hallucinated objects is more pronounced in $\bar{\mathbf{A}}_{T_{p}}^{r}$ and $\bar{\mathbf{A}}_{T_{p}}^{h}$ (typically by a factor of ×1.5–×2.5), indicating that real object tokens exhibit stronger attention to $T_{p}$.}} This suggests that knowledge from $T_{p}$ may support accurate predictions. 

To explain this phenomenon, we re-examine the nature of $V$ and $T_{p}$. 
It is noteworthy that $V$, encoded by the visual encoder, is heterogeneous to the LLM.
Even after fine-tuning efforts to align $V$ with $I_{s}$, an inevitable domain gap persists between them. This gap is widely recognized in many studies~\cite{jiang2024hacl,yu2024hallucidoctor} as a significant cause of hallucinations.
Furthermore, $V$ focuses solely on encoding visual information and is independent of text, thus containing a large amount of instruction-irrelevant visual information. However, during generation, the LLM has access to both the image and user instructions, enabling it to purposefully organize and summarize instruction-related visual information to successfully generate $T_p$. Consequently, $T_p$ inherently encodes instruction-related visual information and naturally belongs to $I_s$.

Therefore, during the generation of each token in $T_p$, the visual content in $V$, which contains substantial instruction-irrelevant information and deviates from $I_s$, is reorganized into $T_p$, which is instruction-relevant and inherently belongs to $I_s$. 
This process produces concentrated and condensed visual features while mitigating the gap between visual features and the text-dominated feature space $I_s$, thereby reducing hallucination and supporting more accurate predictions.
Given this observation, might we exploit the visual feature embedded in $T_p$ to alleviate hallucinations?


\subsection{Increasing Attention to $T_{p}$}
\label{exp}
Motivated by the above insight, we pursued further investigation. To leverage the visual information embedded within $T_{p}$, we propose \textbf{I}ncreased \textbf{A}ttention to $T_{p}$ (IAT). Specifically, for the intermediate layers of the LLM (i.e., layers 5-18), we employ a naive attention amplification mechanism to increase the attention to $T_{p}$:
\begin{equation}
\boldsymbol{\tilde{A}}^{(l,h)}(i)=\boldsymbol{\tilde{A}}^{(l,h)}(i)+\alpha\cdot|\boldsymbol{\tilde{A}}^{(l,h)}(i)|,
\label{eq1}
\end{equation}
where $\alpha$ is the amplification factor, $l \in (5,18)$ and $\mathcal{I}(i) \in T_{p}$. In PAI, $\mathcal{I}(i) \in V$, which is the most critical distinction compared to our approach.

To evaluate the effectiveness of our method, we compare it against PAI, HGAI, and the Greedy decoding. To quantify hallucination rates and linguistic richness in captions, we employ the CHAIR~\cite{chair} and Distinct-1~\cite{distinct-n} metrics. The CHAIR metric quantifies hallucination by verifying whether objects mentioned in captions actually appear in the image, comprising $CHAIR_S$ ($C_S$) and $CHAIR_I$ ($C_I$), which assess sentence-level and instance-level hallucinations respectively:
\begin{equation}
\small
C_S=\frac{\left|\{{captions ~with ~hallucinated ~objects}\}\right|}{\left|\{{all ~captions}\}\right|}
\end{equation}
\begin{equation}
\small
C_I=\frac{\left|\{hallucinated ~objects\}\right|}{\left|\{{all ~mentioned ~objects}\}\right|}
\end{equation}
Distinct-n is a widely used metric for textual diversity assessment, measuring diversity by calculating the proportion of unique n-grams in the generated text:
\begin{equation}
\small
Distinct-n=\frac{unique ~ngram}{total~ngram}
\end{equation}
Specifically, Distinct-1 ($D_1$) evaluates lexical richness and the ability to avoid repetitive patterns in generated text. A higher $D_1$ score indicates richer vocabulary usage and fewer repetitive descriptions.

\begin{figure}[htbp]
    \centering
        \includegraphics[width=1\linewidth]{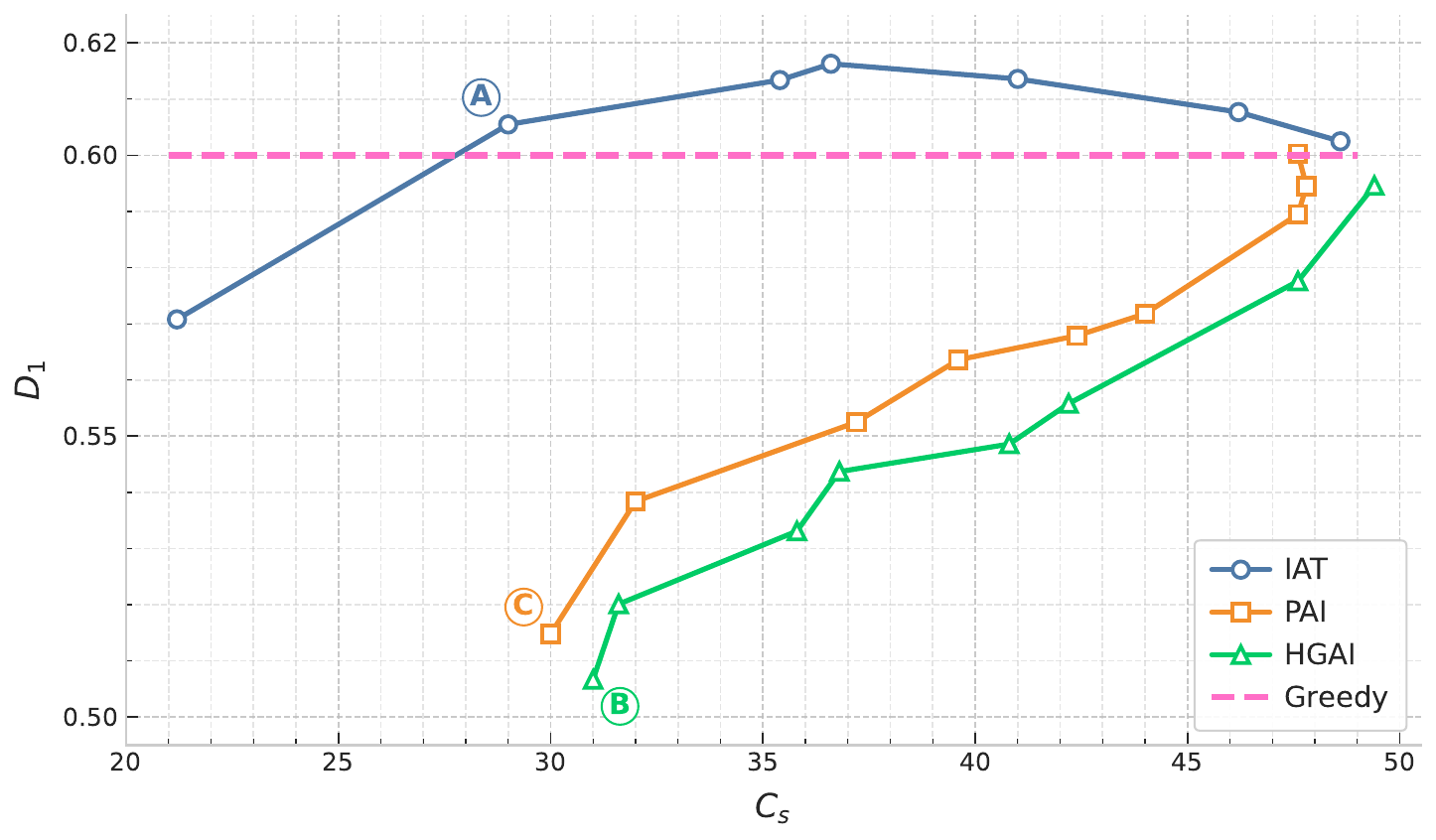}
        \vspace{-15pt}
        \caption{Trends of textual diversity $D_1$ for different methods as the hallucination rate $C_S$ decreases. The dashed line denotes the $D_1$ of the original greedy decoding as a reference. Regions closer to the top-left indicate better performance.}
        \label{fig:plot}
\end{figure}

\begin{figure}[htbp]
    \centering
        \vspace{-10pt}
        \includegraphics[width=1\linewidth]{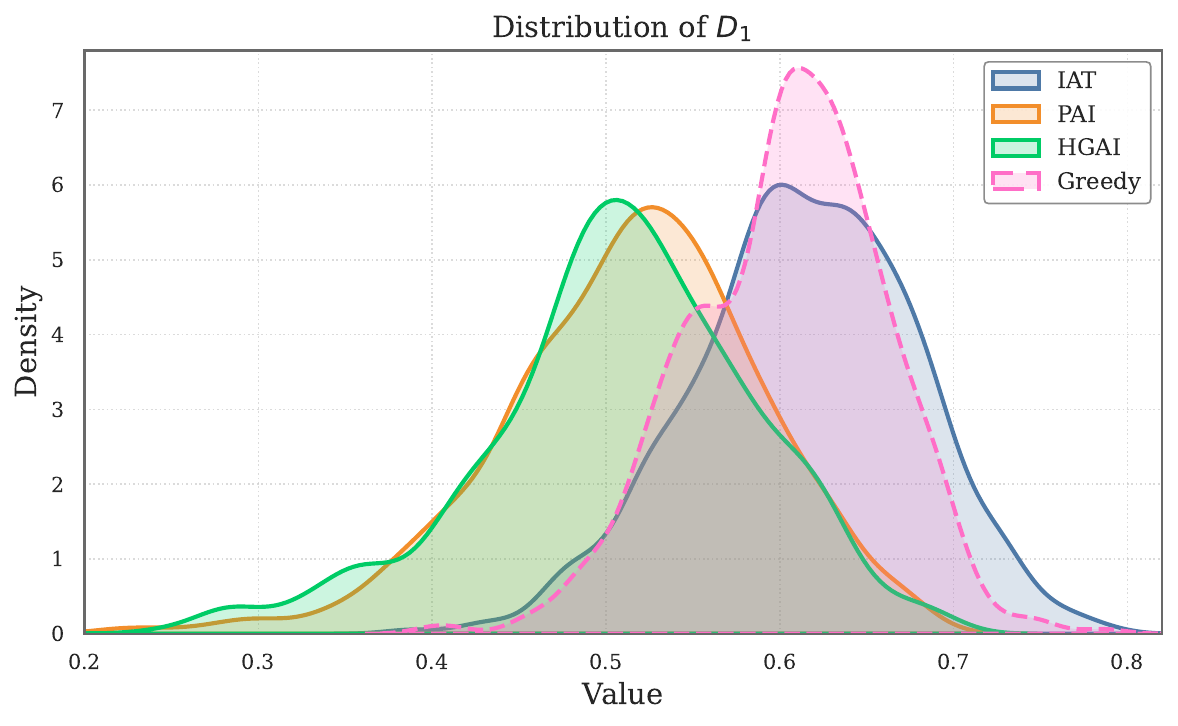}
        \caption{The distributions of textual diversity $D_1$ for captions generated by different methods using the LLaVA-1.5-7B, where a higher $D_1$ corresponds to greater text diversity. The distribution is partially truncated for better visualization.}
        \label{fig:D1_distribution}
        \vspace{-15pt}
\end{figure}

We randomly selected 500 images from COCO~\cite{coco} and generated captions for them using the above four methods on LLaVA-1.5-7B, with the prompt ``Please describe the image in detail.'' For the attention intervention methods, a larger $\alpha$ in Eq.~\ref{eq1} represents a stronger intervention on the model. While this achieves a lower hallucination rate, it inevitably causes greater impairment to certain aspects of the model (e.g, text diversity). Therefore, for PAI, HGAI, and IAT, we varied $\alpha$ and plotted their performance trade-offs between $C_s$ and $D_1$ in Fig.~\ref{fig:plot}, with the $D_1$ of the Greedy decoding annotated for reference. 
It demonstrates that increasing attention to $T_{p}$ effectively mitigates hallucination, providing preliminary validation of our motivation. Additionally, as the hallucination rate decreases, both PAI and HGAI exhibit a noticeable decline in text diversity.

We further selected Greedy and three points (A, B, C) with approximate hallucination rates from Fig.~\ref{fig:plot} and calculated the $D_1$ distributions for their captions and show it in Fig.~\ref{fig:D1_distribution}. We observe that compared to Greedy, PAI and HGAI show a pronounced shift towards lower values, indicating more repetitive descriptions in captions. In contrast, IAT maintains a distribution comparable to Greedy, even exhibiting a higher proportion in the high-score range (0.65-0.8), which suggests that IAT can enhance the diversity of the captions.
This occurs because PAI and HGAI increase attention to all image tokens, which relatively suppresses attention to $T_{p}$. Consequently, the model ``forgets'' previously generated content and tends to repetitively describe the salient objects in the image, leading to redundant language. In contrast, IAT avoids this by increasing attention to $T_{p}$, leveraging contextual knowledge to ensure coherent and diverse text generation.

\begin{figure}[htbp]
    \centering
    \vspace{-10pt}
    \includegraphics[width=0.98\linewidth]{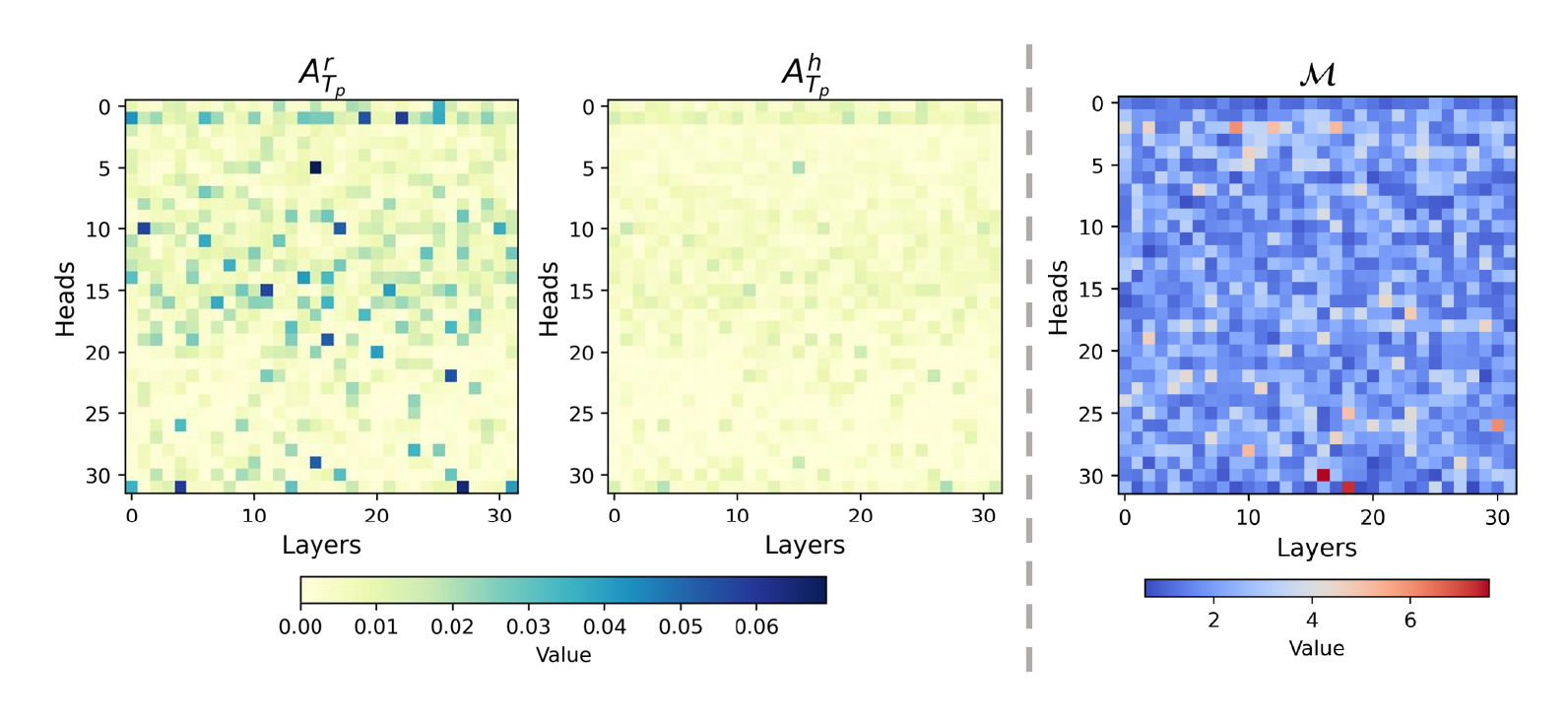}
    \vspace{-13pt}
    \caption{Characteristics of different heads on LLaVA-1.5-7B. Left: Visualization of the average per‑token attention weight to generated text for each layer and head, computed over the sets of real objects (${\mathbf{A}}_{T_{p}}^{r}$) and hallucinated objects (${\mathbf{A}}_{T_{p}}^{h}$). Right: The visualization of ${\mathbf{A}}_{T_{p}}^{r} / {\mathbf{A}}_{T_{p}}^{h}$. Though ${\mathbf{A}}_{T_{p}}^{r}$ is generally higher than ${\mathbf{A}}_{T_{p}}^{h}$, the fold‑difference varies across different heads.}
    \label{fig:heatmap}
\end{figure}

\subsection{Adaptive Attention Enhancement}
While the naive attention amplification method is straightforward and effective, it suffers from several limitations. Therefore, we propose \textbf{Ada}ptive IAT (AdaIAT), adaptively determining both the time to intervene and the amplification magnitude. 
This is achieved by monitoring attention patterns during inference and 
leveraging the different patterns of $\mathbf{A}^{r}$ versus $\mathbf{A}^{h}$ across different attention heads.
AdaIAT is meticulously designed to minimize disruption to the model's standard prediction while simultaneously maintaining potent hallucination mitigation performance.

\begin{table*}[htbp]
  \centering
  \caption{CHAIR hallucination evaluation of various methods on three LVLMs. The best values (excluding Greedy) are highlighted in \textbf{bold}. ${\dagger}$ indicates that the attention intervention is applied based on the Sample decoding.}
    \vspace{-5pt}
  \setlength{\tabcolsep}{1.5mm}{
    \begin{tabular}{l|cccc|cccc|cccc|cccc}
    \toprule
    Models & \multicolumn{4}{c|}{LLaVA-1.5-7B\cite{llava1.5}} & \multicolumn{4}{c|}{LLaVA-1.5-13B\cite{llava1.5}} & \multicolumn{4}{c|}{Janus-Pro-7B\cite{chen2025janus}} & \multicolumn{4}{c}{Qwen2.5-VL-7B\cite{Qwen-VL}} \\
    \hline
    Metrics & $C_S \downarrow$ & $C_I\downarrow$ & F1 $\uparrow$    & $D_1\uparrow$   & $C_S\downarrow$ & $C_I\downarrow$ & F1  $\uparrow$  & $D_1\uparrow$   & $C_S\downarrow$ & $C_I\downarrow$ & F1 $\uparrow$   & $D_1\uparrow$   & $C_S\downarrow$ & $C_I\downarrow$ & F1 $\uparrow$   & $D_1\uparrow$ \\
    \hline
    Sample & 53.0  & 15.6  & 73.4  & 0.65  & 52.0  & 13.7  & 74.9  & 0.66  & 34.0  & 9.0   & 73.5  & 0.64  & 42.0  & 11.3  & 72.5  & 0.67  \\
    AdaIAT$^{\dagger}$ & 42.0  & 12.3  & 73.1  & 0.66  & 37.8  & 10.1  & 75.0  & 0.67  & 29.8  & 8.1   & 72.6  & 0.65  & 34.6  & 9.0   & 75.1  & 0.68  \\
    \hline
    Greedy & 49.0  & 13.3  & 77.9  & 0.60  & 47.8  & 12.5  & 78.9  & 0.60  & 25.8  & 6.7   & 76.8  & 0.62  & 33.6  & 8.3   & 76.5  & 0.65  \\
    PAI\cite{liu2024pai}   & 31.8  & 7.8   & 77.7  & 0.50  & 34.2  & 9.0   & \textbf{79.3}  & 0.56  & 20.4  & 5.6   & 76.1  & 0.61  & 32.0  & 8.2   & 74.7  & 0.64  \\
    HGAI\cite{devil}  & 31.4  & \textbf{6.9}   & 78.3  & 0.50  & 26.4  & 7.8   & 76.9  & 0.55   & 21.0  & 5.3   & 75.9  & 0.62  & 30.4  & 8.0  & \textbf{76.8}  & 0.65  \\
    IAT   & \textbf{29.8}  & 9.0   & 76.8  & \textbf{0.61}  & 31.0  & 8.6   & 78.6  & \textbf{0.61}  & 20.6  & 5.3   & 75.3  & \textbf{0.65} & 29.2  & 7.7   & 76.4  & \textbf{0.67} \\
    AdaIAT & 31.4  & 8.3   & \textbf{79.4}  & 0.60  & \textbf{25.2}  & \textbf{6.1}   & 79.2  & 0.60 & \textbf{19.0} & \textbf{4.9} & \textbf{76.5}  & 0.64  & \textbf{28.4} & \textbf{6.9}   & \textbf{76.8}  & 0.66 \\
    \bottomrule
    \end{tabular}%
    }
  \label{tab:baseline}%
\vspace{-10pt}
\end{table*}%

\subsubsection{Adaptively Determine the Amplification Time}
Naive amplification persistently boosts attention towards $T_{p}$ regardless of whether hallucination tendencies arise. However, hallucinations are sporadic phenomena and LVLMs typically exhibit accurate predictions. Blindly amplifying attention indiscriminately may lead to abnormally high attention on $T_{p}$ during normal predictions, impairing the model's standard prediction capabilities and consequently affecting accuracy. 
An effective amplification strategy should dynamically monitor the attention patterns, promptly detect anomalies (e.g., excessively low attention towards $T_{p}$), and then trigger IAT. Therefore, we establish a layer-wise threshold:
\begin{equation}
\mathcal{T}=\bar{\mathbf{A}}_{T_{p}}^{h} + \beta(\bar{\mathbf{A}}_{T_{p}}^{r}-\bar{\mathbf{A}}_{T_{p}}^{h} )
\end{equation}
where $\beta$ is a balanced coefficient, $\mathcal{T}\in\mathbb{R}^{L}$ and $\mathcal{T}^{(l)}$ denotes the threshold for the $l$-th layer. 
During generation, when $\mathcal{T}^{(l)} > \bar{\mathbf{A}}^{(l)}_{T_{p}}$, it is considered that attention to $T_{p}$ is insufficient, which serves as the trigger for IAT. When $\mathcal{T}^{(l)} \leq \bar{\mathbf{A}}^{(l)}_{T_{p}}$, it is considered that the model maintains adequate attention to $T_{p}$. In this case, the model proceeds with its normal prediction without any intervention, thereby avoiding disruption to its standard prediction behavior.

\subsubsection{Adaptive Amplification Magnitude}
In naive amplification, a fixed $\alpha$ is uniformly applied across all heads. However, compared to real object tokens generation, significant variations exist in different heads when hallucinated 
object tokens are generated. We derive ${\mathbf{A}}_{T_{p}}^{r},{\mathbf{A}}_{T_{p}}^{h} \in \mathbb{R}^{L\times H}$ from datasets $D_{r}$ and $D_{h}$ respectively, which represent the average attention to $T_p$ across heads during real/hallucinated object generation. Subsequently, we compute the attention amplification ratio matrix:
\begin{equation}
\mathcal{M}=\frac{\mathbf{A}^{r}_{T_{p}}}{\mathbf{A}^{h}_{T_{p}}}
\end{equation}
where $\mathcal{M}\in\mathbb{R}^{L\times H}$ and $\mathcal{M}^{(l,h)}$ represents the average attention ratio toward $T_{p}$ for real versus hallucinated object generation at the $h$-th head of the $l$-th layer. We visualize $\mathbf{A}_{T_{p}}^{r}$, $\mathbf{A}_{T_{p}}^{h}$, and $\mathcal{M}$ of LLaVA-1.5-7B in Figure~\ref{fig:heatmap}.
As observed, certain heads exhibit more significant disparities in $\mathcal{M}$, indicating greater attention deficiency toward $T_{p}$ during hallucinated object token generation. Conversely, heads with smaller differences require smaller amplification. Thus, $\mathcal{M}$ serves as a reliable amplification direction, shifting hallucinated attention patterns toward real attention patterns. Consequently, for the $h$-th head in the $l$-th layer, we designate $\mathcal{M}^{(l,h)}$ as its adaptive amplification magnitude and reformulate Equation (5) as follows:
\begin{equation}
\small
\boldsymbol{A}^{(l,h)}(i)=\boldsymbol{A}^{(l,h)}(i)+\alpha\cdot\mathcal{M}^{(l,h)}\cdot\boldsymbol{A}^{(l,h)}(i), \quad
\mathcal{I}(i) \in T_{p}
\label{adaeq}
\end{equation}
\begin{equation}
\small
\boldsymbol{A}^{(l,h)}(k)=\frac{\boldsymbol{A}^{(l,h)}(k)}{\sum_{k}\mathbf{A}^{(l,h)}(k)},\quad
k \in (1,len)
\label{adaeq}
\end{equation}

It is worth noting that as shown in Fig.~\ref{fig:framework}, the naive amplification method intervenes on $\tilde{\mathbf{A}}$, whereas $\mathcal{M}$ represents the ratio of $\mathbf{A}$ for real versus hallucinated object tokens.
Therefore, in Eq.~\ref{adaeq}, we amplify $\mathbf{A}$ and normalize it to ensure $\sum_{i} \mathbf{A}^{(l,h)}(i) = 1$, and $\alpha$ represents the strength of amplification.
In this way, AdaIAT adaptively assigns a specific amplification magnitude for each head. Heads exhibiting larger disparities in $\mathcal{M}$ receive stronger amplification to align with real attention patterns, while those with smaller deviations maintain weaker amplification to avoid disrupting normal attention patterns.


%% file: sec/4_exp.tex
\section{Experiment}



\subsection{Models and Baselines}
We evaluate the effectiveness and generalizability of our method on three representative LVLMs: LLaVA-1.5\cite{llava1.5}, Janus-Pro\cite{chen2025janus}, and Qwen2.5-VL\cite{Qwen-VL}. All models employ the 7B versions, with LLaVA-1.5 additionally employing a 13B version to investigate scale effects.
The number of max tokens for all models is set to 512. Although $\mathcal{T}$ and $\mathcal{M}$ are derived from COCO, they were kept fixed during evaluation on all other datasets. 
We used the Greedy as a reference and selected two representative attention intervention methods, PAI\cite{liu2024pai} and HGAI\cite{devil}, as baselines. Unless specified, all hallucination mitigation methods are implemented based on Greedy decoding.


\subsection{CHAIR Evaluation Results}
The experimental setup for the CHAIR metric follows the configuration of Section~\ref{motivation}, where we use $C_s$ and $C_I$ to evaluate hallucination rates, $D_1$ to assess the textual diversity, and additionally adopt the F1 score to evaluate the richness and accuracy of generated objects, measuring models' prediction capability. Table~\ref{tab:baseline} presents the experimental results of different methods on selected LVLMs. We conducted a hyperparameter search to identify the optimal performance of all methods for reporting. Detailed hyperparameter selections and comparisons with additional methods are provided in the Appendix. It can be observed that for LLaVA-1.5-7B, both PAI and HGAI demonstrate significant hallucination reduction effects. However, this improvement comes at the cost of degraded textual diversity, with $D_1$ decreasing by approximately 15\%. In contrast, IAT and AdaIAT maintain the original text diversity (sustaining $D_1$ around 0.60) of Greedy decoding while achieving comparable low hallucination rates.
This advantage stems from their increased attention to $T_{p}$, thereby preventing repetitive sentences. Moreover, they effectively leverage progressively fusion, instruction‑relevant visual information at each autoregressive step, enabling more accurate predictions with fewer hallucinations.
Furthermore, compared with IAT, AdaIAT adaptively determines when to activate IAT and sets fine-grained amplification magnitudes tailored to each attention head’s characteristics, minimizing disruption to the model’s native prediction patterns. Consequently, under similar hallucination rates and text diversity, AdaIAT outperforms IAT by 2.6 in F1 score, demonstrating superior prediction capability. Similar results are observed on LLaVA-13B.

On more advanced LVLMs like Janus-Pro and Qwen2.5-VL, IAT and AdaIAT exhibit higher $D_1$ scores than PAI and HGAI, demonstrating strengthened linguistic proficiency while maintaining comparable hallucination rate. 
Overall, AdaIAT achieves the lowest $C_s$ and $C_I$ and the highest F1 on Janus-Pro and Qwen2.5-VL while maintaining $D_1$, demonstrating excellent mitigation efficacy across different LVLMs. 
We also integrate AdaIAT with Sample decoding, which effectively alleviates hallucination while maintaining or even improving F1 and $D_1$ scores. This confirms the effectiveness of AdaIAT across different decoding strategies.


\begin{table}[htbp]
\renewcommand{\arraystretch}{0.5}
\setlength{\tabcolsep}{0.018\linewidth} 
  \centering
  \caption{OpenCHAIR hallucination evaluation of different methods on LLaVA-1.5-7B.}
    \begin{tabular}{cccccc}
    \toprule
     & Greedy & PAI\cite{liu2024pai}   & HGAI\cite{devil} & IAT   & AdaIAT \\
    \midrule
    $C_o \downarrow$ & 0.292  & 0.266  & 0.261  & 0.254  & 0.252  \\
    $D_1 \uparrow$    & 0.61  & 0.51  & 0.53  & 0.61  & 0.61  \\
    \bottomrule
    \end{tabular}%
  \label{tab:openchair}%
\end{table}%

\subsection{OpenCHAIR Evaluation Results}
To further validate the effectiveness of our approach, we evaluate OpenCHAIR~\cite{ben2024openchair}, a benchmark that extends the CHAIR metric by relaxing its strong reliance on a closed vocabulary. It first parses generated captions to extract objects, then employs an LLM to judge whether each object is real, hallucinated, or uncertain, and calculates:
\begin{equation}
C_O=\frac{\left|\{{hallucinated ~objects}\}\right|}{\left|\{{hallucinated ~objects ~and ~real ~objects}\}\right|}, 
\end{equation}
We employ LLaVA-1.5-7B (results for other models are detailed in the Appendix) to generate captions for 2,000 images from the OpenCHAIR dataset and summarize the results in Table~\ref{tab:openchair}. It is observed that while PAI and HGAI successfully suppress hallucination rates, they also incur a significant degradation in $D_1$. In contrast, IAT and AdaIAT maintain a $D_1$ score around 0.61 while achieving even lower hallucination rates. This further demonstrates that enhancing attention to $T_{p}$ is effective in mitigating hallucination without compromising language quality.

\begin{figure}[htbp]
    \centering
        \includegraphics[width=0.75\linewidth]{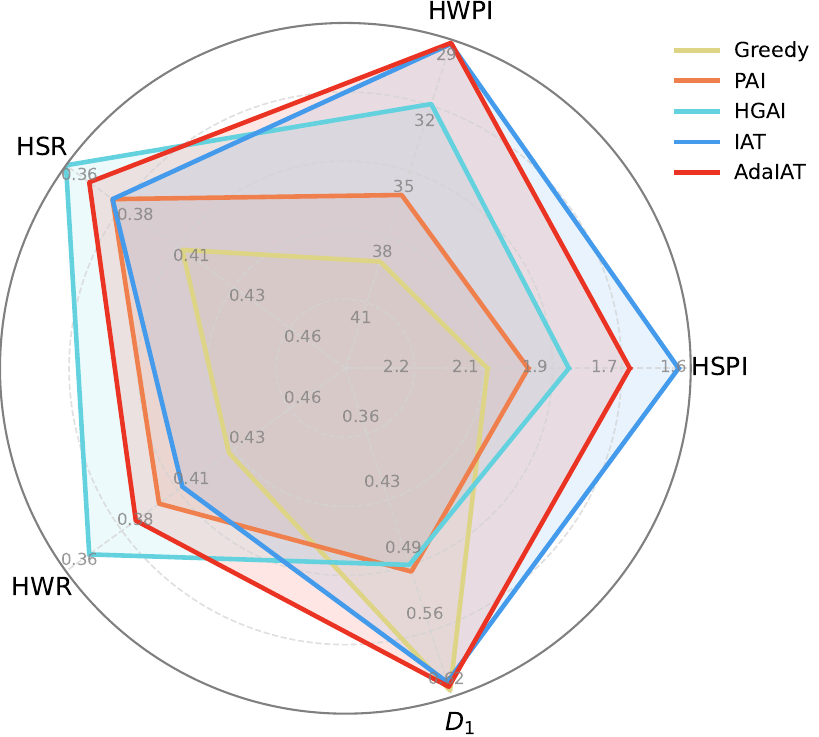}
        \caption{Results of HalluBench in LLaVA-1.5-7B. Five aspects are analyzed: hallucinated sentences per image (HSPI), hallucinated words per image (HWPI), hallucinated sentences ratio (HSR), hallucinated words ratio (HWR), and Distinct-1 ($D_1$). Larger radar area indicates better performance.}
        \label{fig:radar}
\end{figure}

\subsection{GPT-4 Assisted Evaluation}
To more comprehensively evaluate other kinds of hallucinations in captions, such as attributes, positions, and relationships, we further tested the performance of different methods on HalluBench~\cite{HalluBench}.
HalluBench selects 200 images from VG~\cite{VG} and annotates them with detailed object-level description lists. With the assistance of GPT-4, hallucinations in the generated captions are judged sentence by sentence. The experimental results in LLaVA-1.5-7B are shown in Figure~\ref{fig:radar}. It can be observed that AdaIAT significantly outperforms both PAI and HGAI on HWPI and HSPI, achieving improvements of 17\% and 26\% compared to Greedy. While HGAI slightly surpasses AdaIAT on HSR and HWR, it exhibits a significant decline on the $D_1$. Overall, AdaIAT achieves a better balance across all metrics.


\begin{table}[htbp]
\renewcommand{\arraystretch}{0.5}
\setlength{\tabcolsep}{0.018\linewidth} 
  \centering
  \caption{Evaluation results of the text quality for captions generated by different methods on LLaVA-1.5-7B.}
    \begin{tabular}{lccccc}
    \toprule
          & Greedy & PAI\cite{liu2024pai}   & HGAI\cite{devil}  & IAT   & AdaIAT \\
    \midrule
    $D_1 \uparrow$    & 0.614  & 0.498  & 0.492  & 0.609  & 0.610  \\
    $B_{self} \downarrow$ & 0.058  & 0.242  & 0.247  & 0.090  & 0.071  \\
    $B_r \uparrow$  & 85.29  & 84.77  & 84.78  & 85.20  & 85.17  \\
    $B_d \uparrow$  & 59.48  & 56.80  & 56.61  & 58.99  & 58.29  \\
    \bottomrule
    \end{tabular}%
  \label{tab:text_quality}%
\end{table}%

\subsection{Generated Text Quality Evaluation}

To more comprehensively evaluate the quality of captions generated by different methods, we conducted further experiments on the IIW-400~\cite{iiw} dataset, which contains 400 images annotated with highly detailed and hallucination-free captions. We evaluated them using $D_1$ and Self-BLEU~\cite{selfbleu} ($B_{self}$) for language diversity, as well as BertScore\cite{bertscore} for text quality based on the reference captions. $B_r$ and $B_d$ denote the BertScore computed using ``roberta-large'' and ``deberta-xlarge-mnli'', respectively.
The experimental results are summarized in the Table~\ref{tab:text_quality}.

As shown in the results, in terms of $D_1$ and $B_{self}$, both IAT and AdaIAT perform comparably to Greedy. In contrast, PAI and HGAI exhibited significant degradation, revealing the severity of their tendency to produce repetitive text, which substantially harms the linguistic diversity of the model. Similarly, on $B_r$ and $B_d$, PAI and HGAI also demonstrated more pronounced decline. 
Conversely, AdaIAT outperforms HGAI by 0.39 and 1.68 on $B_r$ and $B_d$, respectively, achieving performance similar to Greedy and thus demonstrating well-preserved text quality.

\subsection{Ablations Study}
In this section, we conduct ablation analyses using LLaVA-1.5-7B (results for additional models are provided in the Appendix) to investigate parameter sensitivity (the amplification factor $\alpha$ and the trade-off coefficient $\beta$) and the impact of different layer selections.

\noindent\textbf{Amplification Factor $\alpha$.}
As shown in Tab.~\ref{tab:alpha}, hallucination rates for both IAT and AdaIAT decrease significantly with increasing $\alpha$. However, when $\alpha$ becomes excessively high ($\alpha \geq 0.8$  for IAT or $\alpha \geq 6$ for AdaIAT), the F1 score declines markedly, and $D_1$ also begins to deteriorate. Considering the hallucination rate, predictive capability, and linguistic quality comprehensively, we selected $\alpha=0.8$ for IAT and $\alpha=6$ for AdaIAT as the optimal values.

\begin{table}[htbp]
\renewcommand{\arraystretch}{0.5}
  \centering
  \vspace{-5pt}
  \caption{Performances of IAT and AdaIAT under different amplification factor $\alpha$.}
    \begin{tabular}{lcccccc}
    \toprule
          & $\alpha$ & 0.6   & 0.7   & 0.8   & 0.9   & 1 \\
    \midrule
    \multirow{4}[2]{*}{\begin{sideways}IAT\end{sideways}} & $C_S \downarrow$   & 36.4  & 35.2  & 29.8  & 21.2  & 13.6  \\
          & $C_I \downarrow$    & 10.3  & 10.5  & 9.0   & 7.0   & 5.0  \\
          & F1 $\uparrow$    & 78.1  & 77.4  & 76.8  & 73.5  & 68.4  \\
          & $D_1 \uparrow$    & 0.62  & 0.61  & 0.61  & 0.57  & 0.52  \\
    \midrule
          & $\alpha$ & 4     & 5     & 6     & 7     & 8 \\
    \midrule
    \multirow{4}[2]{*}{\begin{sideways}AdaIAT\end{sideways}} & $C_S \downarrow$    & 34.6  & 33.8  & 31.4  & 31.8  & 29.4  \\
          & $C_I \downarrow$    & 9.5   & 9.3   & 8.3   & 8.5   & 8.0  \\
          & F1 $\uparrow$    & 78.9  & 78.5  & 79.4  & 78.7  & 79.2  \\
          & $D_1 \uparrow$    & 0.60  & 0.60  & 0.60  & 0.59  & 0.59  \\
    \bottomrule
    \end{tabular}%
  \label{tab:alpha}%
\end{table}%

\vspace{-15pt}

\begin{table}[htbp]
\renewcommand{\arraystretch}{0.5}
  \centering
  \caption{Performances of AdaIAT under different balanced coefficient $\beta$.}
    \begin{tabular}{lcccccc}
    \toprule
    $\beta$ & 0.1   & 0.3   & 0.5   & 0.7   & 1 & 1.5 \\
    \midrule
    $C_S \downarrow$    & 32.2  & 32.6  & 31.4  & 33.0    & 32.6 & 34.4 \\
    $C_I \downarrow$    & 9.2  & 8.7  & 8.3  & 8.7   & 8.89 & 9.66 \\
    F1 $\uparrow$    & 79.1 & 79.0 & 79.4 & 78.8 & 78.5 & 78.1  \\
    $D_1 \uparrow$    & 0.599  & 0.600  & 0.601  & 0.597  & 0.599 & 0.592 \\
    \bottomrule
    \end{tabular}%
  \label{tab:adaIAT_beta}%
\end{table}%

\noindent\textbf{Balanced coefficient $\beta$.}
We report the performance of AdaIAT across varying $\beta$ in Tab.~\ref{tab:adaIAT_beta}. Since $\beta$ controls the threshold for activating IAT, a small $\beta$ corresponds to weak intervention, resulting in slight and insignificant hallucination mitigation. 
As $\beta$ increases from 0.1 to 0.5, the hallucination rate continuously decreases. However, an excessively high $\beta$ leads to more frequent activation of IAT, causing abnormally high attention to $T_{p}$, thereby disrupting the model's inherent prediction patterns. 
For instance, when $\beta>0.5$, both $C_S$ and $C_I$ begin to rise, indicating an increase in hallucination rate. Concurrently, F1 score and $D_1$ also start to decline. Therefore, to appropriately control the time of IAT activation, we set $\beta=0.5$.

\begin{table}[htbp]
\renewcommand{\arraystretch}{0.5}
    \setlength{\tabcolsep}{4.8pt}
  \centering
  \vspace{-5pt}
  \caption{The impact of different layer schemes on the hallucination mitigation performance of IAT and AdaIAT.}
    \begin{tabular}{lcccccc}
    \toprule
     & Layers & 0-5   & 18-31 & 0-18  & 5-31  & 5-18 \\
    \midrule
    \multicolumn{1}{c}{\multirow{4}[2]{*}{\begin{sideways}IAT\end{sideways}}} & $C_S \downarrow$    & 24.0  & 26.0  & 4.0   & 1.1   & 29.8  \\
          & $C_I \downarrow$    & 7.1  & 8.1  & 3.3  & 1.8  & 9.0  \\
          & F1 $\uparrow$    & 75.3  & 73.8  & 48.5  & 30.3  & 76.8  \\
          & $D_1 \uparrow$    & 0.582  & 0.604  & 0.163  & 0.036  & 0.606  \\
    \midrule
    \multicolumn{1}{c}{\multirow{4}[2]{*}{\begin{sideways}AdaIAT\end{sideways}}} & $C_S \downarrow$    & 45.2  & 43.8  & 27.4  & 29.0  & 31.4 \\
          & $C_I \downarrow$    & 12.9  & 11.9  & 7.1   & 8.6   & 8.3  \\
          & F1 $\uparrow$    & 77.3  & 78.5  & 77.2  & 78.4  & 79.4  \\
          & $D_1 \uparrow$    & 0.583  & 0.612  & 0.537  & 0.616  & 0.601  \\
    \bottomrule
    \end{tabular}%
  \label{tab:layers}%
\end{table}%

\noindent\textbf{Selection of the Layers to Intervene.}
To further investigate the impact of intervening at the shallow (0–5), intermediate (5–18), and deep(18–31) layers, we designed five different layer selection schemes and report the experimental results in Tab.~\ref{tab:layers}. 
For AdaIAT, intervening solely in shallow or deep layers yielded only marginal hallucination mitigation. Combining intermediate with shallow layers severely degraded textual diversity, whereas adding deep layers led to decreased model accuracy, as reflected in a lower F1 score. 
In contrast, IAT applied only to shallow or deep layers more aggressively reduced hallucinations but at the expense of prediction richness and accuracy (manifested by a reduced F1 score), with the former also sacrificing some textual diversity ($D_1$ dropped by approximately 0.02); combining intermediate with shallow or deep layers caused model collapse.
Therefore, we chose the more balanced-performing intermediate layers for intervention.

\section{Conclusion}
In this work, to address the concern that existing attention intervention methods induce repetitive descriptions, we first analyze attention patterns during real and hallucinated object token generations, revealing that the generated text tokens facilitate real object prediction. Building on this insight, we propose \textit{\textbf{I}ncreased \textbf{A}ttention to Generated \textbf{T}ext} (IAT), thereby significantly reducing the hallucination rate while effectively preserving the model's inherent linguistic capability. To further maintain prediction capability, we propose \textit{\textbf{Ada}ptive IAT} (AdaIAT), which employs a layer-wise adaptive threshold controlling intervention time and fine-grained amplification magnitudes tailored to each attention head's characteristics. Both analysis and experiments demonstrate that the proposed AdaIAT can achieve a good trade-off between hallucination rate, prediction capability, and textual diversity.